\documentclass[conference]{IEEEtran}
\IEEEoverridecommandlockouts
\usepackage{cite}
\usepackage{amsmath,amssymb,amsfonts}
\usepackage{algorithmic}
\usepackage{graphicx}
\usepackage{textcomp}
\usepackage{xcolor}
\usepackage{booktabs}
\usepackage{multirow}
\usepackage{subcaption}
\usepackage{xcolor}
\def\BibTeX{{\rm B\kern-.05em{\sc i\kern-.025em b}\kern-.08em
    T\kern-.1667em\lower.7ex\hbox{E}\kern-.125emX}}

\usepackage{amsmath}

\usepackage{wrapfig}
\usepackage{booktabs}

\usepackage{xurl} 
\usepackage{booktabs}      
\usepackage{graphicx}      
\usepackage{pifont}        
\usepackage{threeparttable}
\usepackage{enumitem}
\usepackage[table]{xcolor}

\definecolor{lightgreen}{RGB}{220,245,220}

\begin{document}






\title{CARE: Contrastive Alignment for ADL Recognition from Event-Triggered Sensor Streams}

\author{%
\makebox[\linewidth][c]{%
\begin{tabular}{@{}c@{\hspace{14mm}}c@{\hspace{14mm}}c@{}}
  \begin{tabular}{c}
    Junhao Zhao$^{*}$\thanks{*This work was done when the author was on an exchange program at UGA.}\\[0.6ex]
    \textit{Electrical and Computer Engineering}\\
    \textit{University of Maryland, College Park}\\
    College Park Maryland, USA\\
    jzhao121@umd.edu
  \end{tabular}
  &
  \begin{tabular}{c}
    Zishuai Liu\\[0.6ex]
    \textit{School of Computing}\\
    \textit{University of Georgia}\\
    Athens, USA\\
    zishuai.liu@uga.edu
  \end{tabular}
  &
  \begin{tabular}{c}
    Ruili Fang\\[0.6ex]
    \textit{School of Computing}\\
    \textit{University of Georgia}\\
    Athens, USA\\
    ruili.fang@uga.edu
  \end{tabular}
  \\[8ex] 

  \multicolumn{3}{c}{%
    \begin{tabular}{@{}c@{\hspace{16mm}}c@{\hspace{16mm}}c@{}}
      \begin{tabular}{c}
        Jin Lu\\[0.6ex]
        \textit{School of Computing}\\
        \textit{University of Georgia}\\
        Athens, USA\\
        jin.lu@uga.edu
      \end{tabular}
      &
      \begin{tabular}{c}
        Linghan Zhang\\[0.6ex]
        \textit{Human Interaction Technology}\\
        \textit{Eindhoven University of Technology}\\
        Eindhoven, Netherlands\\
        l.zhang1@tue.nl
      \end{tabular}
      &
      \begin{tabular}{c}
        Fei Dou$^{\dagger}$\thanks{$^{\dagger}$Corresponding author}\\[0.6ex]
        \textit{School of Computing}\\
        \textit{University of Georgia}\\
        Athens, USA\\
        fei.dou@uga.edu
      \end{tabular}
    \end{tabular}
  }
\end{tabular}%
}
}


\maketitle

\begin{abstract}
The recognition of Activities of Daily Living (ADLs) from event-triggered ambient sensors is an essential task in Ambient Assisted Living, yet existing methods remain constrained by representation-level limitations. Sequence-based approaches preserve temporal order of sensor activations but are sensitive to noise and lack spatial awareness, while image-based approaches capture global patterns and implicit spatial correlations but compress fine-grained temporal dynamics and distort sensor layouts. Naïve fusion (e.g., feature concatenation) fails to enforce alignment between sequence- and image-based representation views, underutilizing their complementary strengths. We propose
\underline{C}ontrastive \underline{A}lignment for ADL \underline{R}ecognition from \underline{E}vent-Triggered Sensor Streams (CARE), an end-to-end framework that jointly optimizes representation learning via Sequence–Image Contrastive Alignment (SICA) and classification via cross-entropy, ensuring both cross-representation alignment and task-specific discriminability. CARE integrates (i) time-aware, noise-resilient sequence encoding with (ii) spatially-informed and frequency-sensitive image representations, and employs (iii) a joint contrastive-classification objective for end-to-end learning of aligned and discriminative embeddings. Evaluated on three CASAS datasets, CARE achieves state-of-the-art performance (89.8\% on Milan, 88.9\% on Cairo, and 73.3\% on Kyoto7) and demonstrates robustness to sensor malfunctions and layout variability, highlighting its potential for reliable ADL recognition in smart homes. {We release our code at \url{https://github.com/Jhziiiig/CARE}.}
\end{abstract}


\section{Introduction}

Global increases in life expectancy are leading to aging societies, with a rising number of older adults who require continuous support from healthcare providers and their family members~\cite{who2022}. However, given the critical shortage of healthcare personnel, it is essential to support older adults in maintaining independence for as long as possible~\cite{medshortage}. An essential component of independent living is the ability to perform Activities of Daily Living (ADLs)---such as toileting, dressing, feeding, and cooking---that directly affect an individual’s quality of life. These functional abilities often decline with aging and can be further compromised by age-related chronic conditions~\cite{prince2013global}. 

Ambient Assisted Living (AAL) technologies have emerged to support ADL performance, encompassing systems for activity recognition, anomaly detection, and personalized prompting. Among these, \textit{ambient sensor-based systems} are especially attractive for AAL, as they provide unobtrusive and privacy-friendly monitoring without wearables' discomfort or cameras' intrusiveness~\cite{dunne2021survey}. Embedded in home environments, ambient sensors offer passive, continuous, and cost-effective monitoring of motion, door usage, and environmental states, enabling scalable deployment in real-world homes
~\cite{cook2012casas}.

Despite these advantages, ADL recognition from ambient sensors is inherently challenging. Different from continuously sampled wearable sensor signals, ambient sensors generate \textit{event-triggered time series} that are sparse, irregular, and noisy: motion sensors activate only when residents pass by, door sensors only when opened or closed. Moreover, older adults often perform the same ADL with variations in order, speed, or completion, further complicating recognition~\cite{weakley2019naturalistic}. To be reliable, recognition models must capture fine-grained temporal dynamics while also contextualizing spatial information across multiple sensors~\cite{hossain2018deactive, hiremath2022bootstrapping, arrotta2022dexar}.  

\setlength{\tabcolsep}{1mm} %
\begin{table*}[tb]
  \centering
  \caption{Comparison of Encoding Methods for Event-Triggered Time Series in ADL Recognition.}
  \label{tab:work_extended}
  \resizebox{0.8\textwidth}{!}{%
  \begin{tabular}{l|cccccccccc}
    \toprule
    Model & Input Repr. & Encoder & View & TimeResolution & PSI* & NR \dag & Fusion Strat. & LO \ddag  & End2End & Dataset \\
    \midrule
    DeepCASAS~\cite{liciotti2020sequential} & Sequence & BiL & Uni-View &  \ding{55}         & \ding{55} & \ding{55} & None              & CE   & Yes     & CASAS \\
    JointTemporalModel\cite{app10155293} & Sequence & BiL+1DCNN & Uni-View & Minute & \ding{55} & \ding{51}& None & CE & Yes & Ordonez, Kastern \\
    EmbeddingFCN~\cite{bouchabou2021fully}  & Sequence & FCN  & Uni-View & \ding{55}         & \ding{55} & \ding{55} & None              & CE  & Yes      & CASAS \\
    TCN~\cite{lea2016temporal}           & Sequence & TCN & Uni-View & Hourly         & \ding{55} & \ding{55} & None              & CE   & Yes     & Self-conducted \\
    \midrule
    BinaryCNN~\cite{9245533} & Image & CNN & Uni-View & \ding{55} & \ding{55}& \ding{55} & None & CE & Yes & SH \\
    DCNN~\cite{gochoo2018unobtrusive}       & Image  & CNN   & Uni-View & \ding{55}         & implicit & \ding{55} & None              & CE   & Yes    & CASAS \\
    GreyDCNN~\cite{mohmed2020employing}     & Image  & CNN   & Uni-View & Local ($\Delta t$)      & implicit & \ding{55} & None              & CE     & Yes    & Smart Home \\
    \midrule
    GraphModel~\cite{s24123944}             & Graph  & GNN  & Uni-View  & Unit Duration  & implicit & Partial   & None              & CE     & Partial    & CASAS \\
    \midrule
    TDOST~\cite{TDOST}                      & Textual & BiL+SE{$\star$}  & Uni-View & Begin+Interval & description & \ding{55} & None              & CE   & Partial    & CASAS, Orange \\
    \midrule
    \textbf{Ours (CARE)}                    & Seq+Img   & BiL+Res  & Cross-View & Hour-binned    & \ding{51} & \ding{51} & ContraAlign & SICA+CE  & Yes  & CASAS \\
    \bottomrule
  \end{tabular}
  }
  \\[2pt]
  \noindent\footnotesize\emph{*PSI:} Physical Spatial Information, \dag \emph{NR}: Noise Robust, \ddag \emph{LO}: Learning Objective, $\star$ \emph{SE}: Sentence Encoder, \emph{BiL}: BiLSTM, \emph{CE}: Cross-Entropy.
\end{table*}

Two dominant strategies have been explored for encoding event-triggered ADL data. \textit{Sequence-based methods} ~\cite{liciotti2020sequential,bouchabou2021fully, app10155293, lea2016temporal}  preserve event order to capture temporal dependencies, thereby modeling daily routines. However, they often ignore the absolute daily time context (e.g., morning vs. evening), are highly sensitive to noise, and lack explicit mechanisms for capturing spatial relationships among sensors. \textit{Image-based methods}~\cite{9245533,gochoo2018unobtrusive,mohmed2020employing}, in contrast, transform event streams into activity images, thereby leveraging global patterns and the implicit spatial correlations embedded in sensor indices. Yet, they suffer from distorted layouts introduced by artificial 1D indexing, reduced temporal resolution that obscures fine-grained dynamics, and the neglect of sensor activation frequency as a potentially discriminative feature.

These limitations highlight a major gap: existing methods focus exclusively on either sequence or image encodings, treating temporal and spatial cues in isolation, and missing the opportunity to exploit their natural complementarity. Prior work has explored simple fusion strategies—such as feature concatenation or late fusion.  While naïve fusion strategy (simple concatenation) could, in principle, combine both views, it ignores cross-view interactions and lacks explicit constraints to ensure that both representation views are effectively utilized, limiting its ability to fully leverage complementary information. To overcome this, we argue for a unified framework that can jointly leverage both sequence and image perspectives within a shared latent space.

To this end, we propose \textbf{CARE} (A \underline{C}ontrastive \underline{A}lignment  Framework for ADL \underline{R}ecognition from  \underline{E}vent-Triggered Sensor Streams),  an end-to-end framework that unifies sequence- and image-based representations through supervised contrastive alignment and joint classification learning. 
{CARE is novel at the \emph{framework} level: it treats sequence and image as two complementary but heterogeneous views of the same event stream, and introduces an explicit \emph{cross-view alignment principle} (SICA: \underline{S}equence–\underline{I}mage \underline{C}ontrastive \underline{A}lignment) to learn a shared latent space that is simultaneously view-consistent and class-discriminative, jointly optimized with task supervision in one stage.}


\noindent {The main contributions of this work are:}  
\begin{itemize}[leftmargin=*]
\item {\textit{A cross-view alignment framework for event-triggered ADL recognition (CARE).} We propose a backbone-agnostic framework that aligns sequence and image views of the same event stream, and jointly optimizes cross-view alignment and task supervision in one stage.}
\item \textit{Sequence--Image Contrastive Alignment (SICA).} We introduce a supervised contrastive alignment objective that unifies sequence- and image-based embeddings in a shared latent space, enforcing view-invariant and class-consistent representations.
\item \textit{Robust temporal encoding.} We design a time-aware, noise-resilient sequence encoding that preserves global daily context while suppressing spurious sensor firings, addressing the limitations of prior order-only encodings.
\item \textit{Spatially-informed and frequency-sensitive image representation.} We propose a dual-perspective image representation that integrates temporal patterns with sensor-layout spatial cues and frequency-sensitive transitions, yielding richer and more interpretable activity signatures.

\end{itemize}



\begin{figure*}[htbp]
  \centering
   \includegraphics[width=0.95\linewidth]{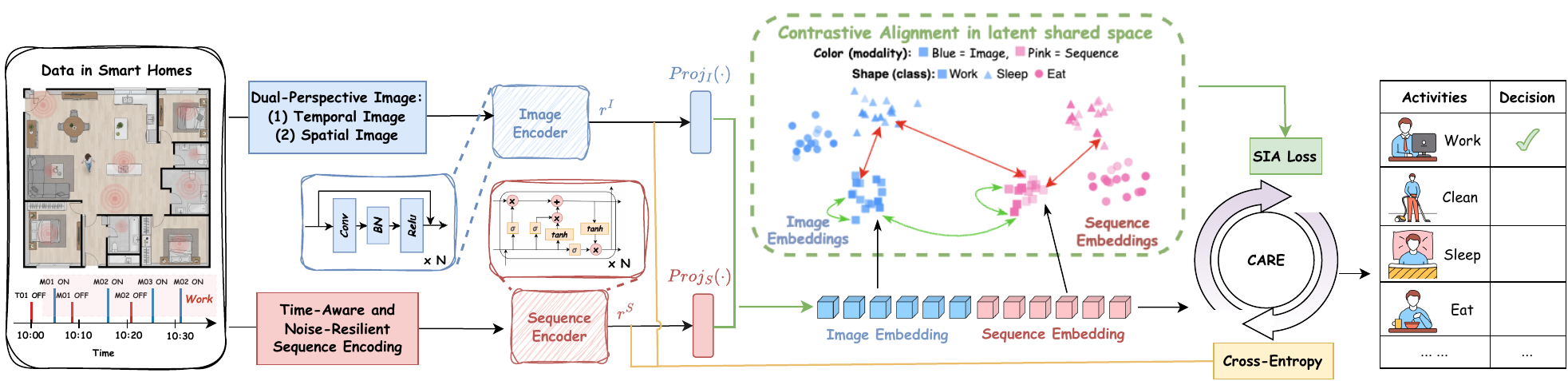}
   \caption{\underline{C}ontrastive \underline{A}lignment for ADL \underline{R}ecognition from \underline{E}vent-Triggered Sensor Streams (CARE) Framework.}
   \label{fig:framework_ver1}
\end{figure*}

\section{Related Work}

The integration of ambient sensors into assistive living technologies has shown strong potential to promote independent living and enable early detection of health deterioration~\cite{manoj2018active}. 
Research has demonstrated that ambient sensor data, when analyzed with advanced machine learning techniques, can effectively classify and predict ADLs with high accuracy~\cite{acampora2013survey}. 
Ambient sensor-based systems are especially attractive for AAL because they are unobtrusive, cost-effective, and scalable, offering continuous monitoring of motion, door usage, and environmental states in naturalistic homes~\cite{cook2012casas}.
Moreover, they align with broader trends toward privacy-preserving monitoring, respecting autonomy while enabling timely health alerts~\cite{ghosh2023ultrasense}. However, the event-triggered nature of such data makes ADL recognition inherently challenging: signals are sparse, irregular, and noisy, and older adults often perform the same activity with variations in order, speed, or completion~\cite{weakley2019naturalistic}. These characteristics underscore the need for robust representation learning methods that can jointly capture fine-grained temporal dynamics and cross-sensor spatial context~\cite{hossain2018deactive, hiremath2022bootstrapping, arrotta2022dexar}.
Table~\ref{tab:work_extended} summarizes representative encoding strategies of event-triggered sensor data for ADL recognition. 
Below, we review four main lines of work.

\noindent \textbf{Sequence-based Deep Learning Methods.}  
A widely adopted approach is to model event-triggered sensor streams as temporal sequences, leveraging deep sequence models to capture event order and temporal dependencies. LSTM and its variants (e.g., Bi-LSTM) have been applied extensively for activity recognition~\cite{oguntala2021passive, liciotti2020sequential}, with extensions that integrate convolutional layers to enhance local feature extraction~\cite{app10155293}. Other architectures, such as Fully Convolutional Networks (FCN)~\cite{bouchabou2021fully} and Temporal Convolutional Networks (TCN)~\cite{lea2016temporal} have also been explored. These models, originally popularized in domains like natural language processing, enable the learning of hierarchical dependencies and activity flow from sensor event sequences.  
While effective at capturing the temporal flow of activities, these methods often \textit{ignore absolute daily time context}, which is crucial for distinguishing activities that share similar local patterns but occur at different times (e.g., sleeping at night vs.\ napping in the afternoon). Moreover, sequence-only methods are highly sensitive to noise and outliers~\cite{doi:10.1177/14759217221098569}, and they fail to explicitly capture cross-sensor spatial relationships that are key to disambiguating similar activities.

\noindent \textbf{Image-Based Encoding Approaches.}  
An alternative line of work transforms event streams into images and applies convolutional encoders. BinaryCNN~\cite{9245533}, DCNN~\cite{gochoo2018unobtrusive} and GreyDCNN~\cite{mohmed2020employing} map sensor activations into binary or grayscale activity images, allowing CNNs to exploit global structures and long-range correlations of time series. These approaches are generally more robust to noise than sequence encoders, since image representations smooth over missing or spurious events. However, several limitations remain: (i) the commonly used 1D sensor indexing distorts the true 2D spatial layout, placing unrelated sensors adjacently and losing room-level semantics; (ii) temporal details are compressed into a single axis or pixel intensity, weakening fine-grained activity dynamics; and (iii) all sensor activations are treated equally, overlooking activation frequency as a discriminative cue.

\noindent \textbf{Other Input Representations.}  
Graph-based approaches~\cite{s24123944} represent sensor activations as nodes and transitions as edges, enabling explicit modeling of cross-sensor relationships. While effective on certain datasets, such models are computationally expensive due to the repeated graph construction and inference iterations, which limit their scalability in real-world deployments. More recently, textual encodings have emerged: Jüttner et al.~\cite{JüttnerFleigBuchmann+2025+173+187} applied GPT-4 to infer ADLs and detect anomalies in CASAS~\cite{cook2012casas}, while TDOST~\cite{TDOST} tokenizes event sequences into words and leverages a pretrained sentence transformer followed by a BiLSTM and classifier. Notably, TDOST achieves state-of-the-art performance across several benchmarks without explicit temporal features, underscoring the potential of language-inspired representations. However, when this method plugs the time information into the words, the performance is not improved. This highlights an open question: how to exploit temporal information in non-traditional encodings like text while preserving underlying activity patterns.

\noindent \textbf{Contrastive Learning for ADL Recognition.}  {Contrastive learning learns representations by bringing semantically similar samples closer and pushing dissimilar ones apart in the embedding space \cite{chen2020simple}. 
It has achieved strong results in vision and NLP~\cite{khosla2020supervised}, and has increasingly been explored for human activity recognition (HAR). Representative examples include CPC~\cite{haresamudram2021contrastive}, which applies contrastive objectives to learn temporal features, and multi-view HAR frameworks such as MuJo~\cite{fritsch2025mujo} and SEAL~\cite{ge2025semantically}, which align representations across modalities or between sensor signals and label semantics. However, these studies primarily target continuously sampled wearable or rich multimodal streams (e.g., IMU, video, pose, text), and do not directly address \emph{event-triggered} ambient sensor streams that are sparse, irregular, and noisy. As a result, the adoption of contrastive learning for ambient, event-triggered ADL recognition remains limited~\cite{chen2024enhancing, chen2024utilizing}, where most prior methods still adopt a \emph{single-view} encoding (sequence-only or image-only), or use \emph{naïve fusion} (feature concatenation with cross-entropy) when multiple views are available. Such designs provide no explicit mechanism to align heterogeneous views, and may underutilize complementary temporal and spatial cues. 
CARE fills this gap by introducing a supervised cross-view contrastive objective to align sequence and image embeddings in a shared latent space.}

\section{Proposed Method}


Raw event-triggered sensor streams are first processed through a preprocessing pipeline, including segmentation, index mapping, padding, and signal normalization, to generate both image-based and sequence-based input representations.  
The sequence branch further processes the streams with temporal binning and frequency-based event filtering, followed by a recurrent encoder that captures temporal correlations and noise-resilient patterns. The image branch
generates dual-perspective image representations consisting of a temporal image and a spatial image, which are encoded by convolutional networks to model spatial dependencies and global patterns. The encoders' outputs are projected into a shared latent space, where a contrastive alignment objective enforces view-invariant, class-consistent embeddings. Along with a cross-entropy classification head, the proposed CARE framework (shown in Fig.~\ref{fig:framework_ver1}) jointly performs robust representation learning and end-to-end activity recognition. 


\subsection{Time-Aware and Noise-Resilient Sequence Encoding}
\label{sec:seq}

The aim of sequence encoding is to transform irregular and noisy event-triggered sensor streams into structured temporal representations that capture daily contextual 
routines while remaining robust to noise activations. Such representations should capture when activities are likely to occur and how sensor activations evolve over time,  providing a foundation for downstream ADL recognition tasks. 

\noindent \textbf{Temporal Binning.} 
Prior sequence encodings often struggle to handle temporal information. Existing methods\cite{app10155293,lea2016temporal,mohmed2020employing,s24123944,TDOST} typically:  
(i) ignore timestamps entirely, discarding daily context; (ii) use inter-event intervals $\Delta t_i = t_i - t_{i-1}$, which capture local timing gaps but fail to reflect absolute daily temporal positioning; or (iii) combine inter-event intervals with activity begin times $(t_{\mathrm{start}}, \Delta t_i)$, which anchors the sequence to its onset and preserves absolute time-of-day information at the activity level, but does not explicitly associate each sensor event with its own global timestamp. 
In contrast, our method assigns a coarse-grained temporal embedding to every event. 
For each raw time-of-day $t_i \in \mathbb{R}^+$, we compute
\[
\tau_i = \left\lfloor \frac{t_i}{\Delta T} \right\rfloor,
\]
where $\Delta T$ is a dataset-specific bin width (e.g., one hour for milan). This design ensures that each sensor event is embedded within the daily routine context (e.g., nighttime vs. midday) while suppressing minute-level fluctuations that are irrelevant for hour-level ADL recognition. The discretized index $\tau_i$ is then concatenated with sensor ID and signal state to form a event representation.

\noindent \textbf{Frequency-based Event Filtering.}
Event-triggered sequences often contain accidental  sensor activations that appear only once or a few times within an activity, yet unnecessarily inflate the sequence length and introduce noise.  Naïve strategies, such as retaining all events, preserve irrelevant noise that can degrade recognition performance.  

\begin{figure}[tb]
  \centering
   \includegraphics[width=0.55\linewidth]{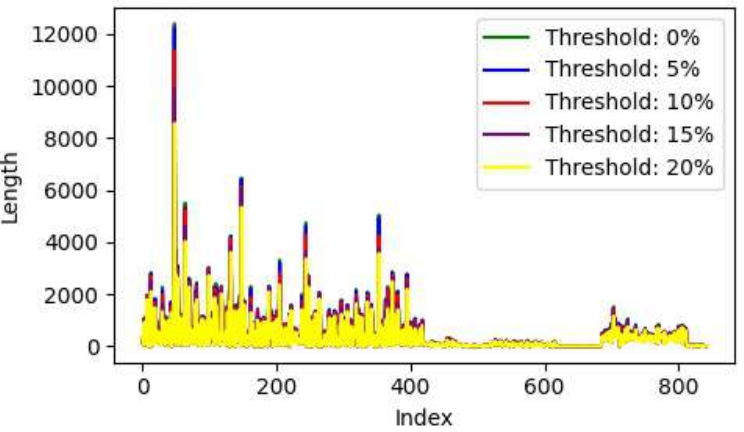}
\caption{Frequency-based Event Filtering on \textit{Cairo}.}
\label{fig:freq}
\end{figure}

To achieve robustness, we adopt a frequency-based criterion to remove unreliable events. For an activity segment with $n$ events, let $f_j$ denote the activation count of sensor $s_j$. We normalize these counts as
$
\tilde{f}_j = \frac{f_j}{\max_{k} f_k},
$
so that the most frequently activated sensor has $\tilde{f}_{j_{max}} = 1$.  
A sensor event is retained only if
$
\tilde{f}_j \geq \theta,
$
where $\theta \in (0,1)$ is a dataset-specific threshold.  
As illustrated in Fig.~\ref{fig:freq}, 
increasing the threshold $\theta$ progressively prunes low-frequency activations and shortens the sequences, trading off noise reduction against potential loss of rare but informative events.

\begin{figure}[tb]
  \centering
   \includegraphics[width=0.98\linewidth]{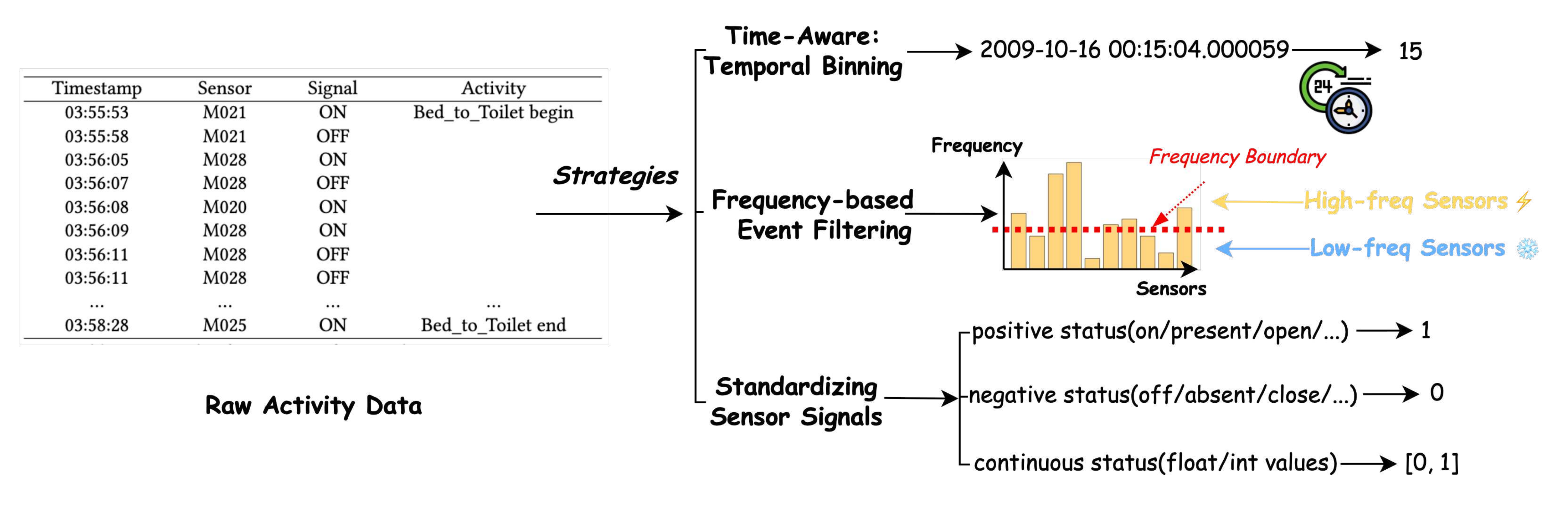}
   \caption{Sequence Representation.}
   \label{fig:seq_encoding}
\end{figure}

\noindent \textbf{Sequence Representation.} To ensure consistency across heterogeneous sensor types, binary states (e.g., ``ON'', ``OPEN'', ``PRESENT'') are mapped to 1 and others to 0, while continuous values such as temperature (e.g., 20.5 and 19.5) are normalized to $[0,1]$. Each activity segment is further padded or truncated to a fixed length $L$ to enable batch processing. Each event is then represented by concatenating a one-hot encoding of its sensor ID, 
the temporal index $\tau_i$, and the normalized signal state. The resulting enriched sequence is processed by a sequence encoder (e.g., LSTM) to generate temporal embeddings $h_t$ that jointly capture sensor identity and 
contextualized temporal dependencies (Fig.~\ref{fig:seq_encoding}).

\subsection{Spatially-Informed and Frequency-Sensitive Image Representation }



While sequence encoding captures temporal correlations, it is limited in modeling 
global structures across long sequences and inter-sensor relationships. An alternative 
approach is to transform event-triggered sequences into 2D activity images, allowing 
CNNs to exploit spatial patterns and long-range dependencies in a structured way.

Prior studies~\cite{gochoo2018unobtrusive,mohmed2020employing} typically represent 
activities as binary or grayscale matrices, with the $x$-axis denoting event order and 
the $y$-axis denoting sensor IDs. This design has several advantages: it preserves the 
temporal order of events and, to some extent, encodes implicit spatial relations through 
the vertical sensor index. However, the spatial information is inaccurate, as the 
one-dimensional $y$-axis does not reflect the true two-dimensional floor-plan. Sensors 
located in different rooms may be placed adjacently in the $y$-axis, leading to 
semantic confusion. Furthermore, such encodings treat all sensor activations equally 
and ignore their activation frequencies, failing to distinguish frequent transitions 
from sporadic noise. 

To overcome these limitations, we propose a \textit{dual-perspective image representation} consisting of a temporal image and a spatial image:  

\noindent \textbf{Temporal Image.} 
For an activity segment with events $\{e_i\}_{i=1}^n$, where $e_i=(\tau_i, s_i, a_i)$ denotes temporal index, sensor ID, and normalized activation signal, we construct a temporal image $I^{\mathrm{temp}}$ such that
$
I^{\mathrm{temp}}(x_i,y_i) = \phi(\mathrm{idx}(\tau_i), s_i, a_i),
$
where $x_i$ corresponds to the event order ($x_1$ denotes the first 
timestamp, followed sequentially up to $x_L$ after padding or truncation), $y_i$ 
is the index of sensor $s_i$, and $\mathrm{idx}(\tau_i)$ is the 
discretized time index of timestamp. The mapping function 
$\phi(\cdot)$ integrates the time index, sensor identity, and signal state into color channels (e.g., motion = blue, door = yellow, temperature = red). This representation highlights temporal dynamics while differentiating sensor modalities. 

\noindent \textbf{Spatial Image.} Each sensor $s_j$ is associated with normalized floor-plan coordinates 
$(u_j,v_j) \in [0,256)^2$. We construct a spatial image $I^{\mathrm{spat}}$ where 
nodes represent sensor positions and edges represent transitions between consecutive 
events. Nodes are rendered according to their activation frequency $f_j$ (already defined 
in Sec.~\ref{sec:seq}), normalized by $\max_k f_k$, with darker colors indicating 
more frequently activated sensors. For edges, the transition from $s_j$ to $s_k$ is assigned weight
\[
w_{jk} = {c_{jk}} / {\max_{p,q} c_{pq}},
\]
where $c_{jk}$ is the number of observed transitions between $s_j$ and $s_k$ in 
the segment, with darker edges denoting more frequent transitions. States such as 
``OFF'', ``CLOSE'', and ``ABSENT'' are excluded to reduce clutter. This representation leverages the floor-plan to embed spatial context and highlights recurrent patterns while suppressing sporadic noise.  

\begin{figure}[tb]
  \centering
   \includegraphics[width=0.95\linewidth]{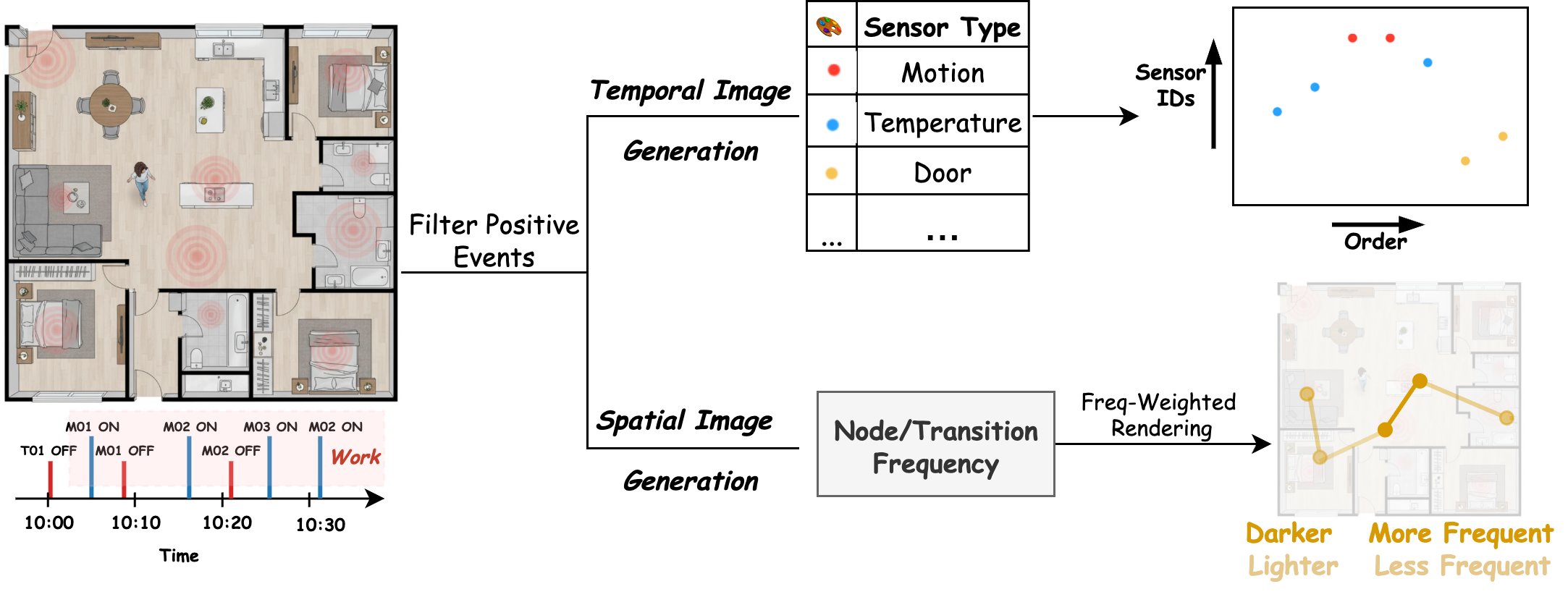}
   \caption{Image Representation.}
   \label{fig:image_encoding}
\end{figure}

\noindent \textbf{Unified Image Representation.}  
The temporal and spatial images are concatenated to form a composite image
$
I^{\mathrm{comp}} = \text{concat}(I^{\mathrm{temp}}, I^{\mathrm{spat}}),
$
which jointly encodes temporal ordering, spatial configuration, and frequency awareness. 
The composite image is then processed by a CNN encoder to produce an embedding that is 
spatially-informed, frequency-sensitive, and robust to noise 
(Fig.~\ref{fig:image_encoding}). { The floor-plan background is shown for illustration only and is not included in spatial images.}
\label{sec:img}

\subsection{Sequence–Image Contrastive Alignment (SICA)}
\label{sec:scl}

Our goal is to unify the complementary strengths of the sequence- and image-domain streams by learning a shared latent space. The \textbf{sequence-domain stream} captures temporal dependencies explicitly. The \textbf{image-domain stream} emphasizes spatial relationships and frequency-sensitive patterns, and preserves implicit temporal order via event sequencing. These two perspectives are complementary and partially overlapping: one models fine-grained temporal dynamics, the other grounds activities in spatial context with implicit temporal correlations.

Without explicit interaction, however, training the two streams independently produces heterogeneous embeddings that cannot be directly fused, limiting cross-view synergy. A straightforward alternative is to concatenate the outputs of the two encoders and optimize a cross-entropy loss. While simple, this naive fusion does not explicitly enforce alignment between views: the classifier may overfit to the dominant stream, ignore complementary cues, and fail to enforce consistency between sequence- and image-based embeddings of the same activity.  

To address these limitations, we propose \textbf{Sequence–Image Contrastive Alignment (SICA)}, which introduces a contrastive objective (i.e., the SICA loss) to explicitly align sequence and image embeddings in a shared latent space. By pulling together cross-view embeddings of the same activity and pushing apart those of different activities, SICA enforces cross-view alignment and class-level discriminability, enabling the model to exploit temporal and spatially grounded cues.

\noindent \textbf{Projection Heads.}  
To enable contrastive alignment across heterogeneous representation views, we introduce lightweight projection heads after sequence and image encoders. Each projection head is a single-layer MLP that maps view-specific encoder outputs into a shared latent space with matched dimensionality. This transformation ensures comparability across views and refines representations for contrastive learning. 

\noindent \textbf{Positive and Negative Pair Construction.}  
Given a mini-batch $B$ of activity segments, each input $\mathbf{x}_i$ produces two embeddings after the projection heads: a sequence embedding $\mathbf{z}_i^S$ and an image embedding $\mathbf{z}_i^I$. For each anchor embedding, positives are defined at two levels:  

\noindent - \emph{Intra-view positives}: embeddings from other samples of the same class within the same view of encoding (e.g., $(\mathbf{z}_i^S$, $\mathbf{z}_j^S)$ if $y_j = y_i$, where $y$ represents class label.).  

\noindent - \emph{Cross-view positives}: the paired embedding of the same sample across 
domains ($\mathbf{z}_i^S$ vs. $\mathbf{z}_i^I$), together with embeddings from the same class in the alternate view (e.g., $(\mathbf{z}_i^S$,$\mathbf{z}_j^I)$ if $y_j = y_i$).  

All embeddings from different ADL classes, regardless of view, are treated as negatives. This design enforces both cross-view alignment and class-level cohesion, while ensuring separation between different activities. 

\noindent \textbf{Contrastive Objective.} Let $P_{in}(i)$ and $P_{cr}(i)$ denote the intra-view and cross-view positive sets for anchor $i$, and let $A(i)$ denote the set of all non-anchor embeddings in the batch. The instance-level loss is defined as:

\begingroup
\small
\begin{equation}
\label{eq:sica}
\begin{aligned}
\mathcal{L}_{\mathrm{SICA},i}
&= \frac{-1}{|P_{in}(i)| + |P_{cr}(i)|}\Bigg(
   \sum_{p\in P_{in}(i)}\!\! \log\frac{\exp(\langle \mathbf z_i,\mathbf z_p\rangle/\tau)}
   {\sum_{a \in A(i)} \exp(\langle \mathbf z_i,\mathbf z_a\rangle/\tau)} \\
&\qquad\quad
 + \sum_{p\in P_{cr}(i)} \log\frac{\exp(\langle \mathbf z_i,\mathbf z_p\rangle/\tau)}
   {\sum_{a \in A(i)} \exp(\langle \mathbf z_i,\mathbf z_a\rangle/\tau)}
\Bigg),
\end{aligned}
\end{equation}
\endgroup
where $\langle \cdot,\cdot\rangle$ denotes cosine similarity and $\tau$ is a 
temperature hyperparameter. The overall batch loss is
\[
\mathcal{L}_{\mathrm{SICA}} = \tfrac{1}{2|B|}\sum\nolimits_{i} \mathcal{L}_{\mathrm{SICA},i}
\]

This formulation enforces that embeddings of the same activity—across both representation views—are closely aligned, while embeddings of different activities remain well separated. By integrating intra- and cross-view positives, SICA achieves cross-view alignment and class-consistent coherence, producing a shared latent space that is both discriminative and robust.

\bigskip

\subsection{CARE: An End-to-End Framework of Contrastive Alignment for ADL Recognition}

While SICA achieves representation-level alignment by enforcing intra-class cohesion and cross-view consistency, it requires a second-stage classifier to be trained on top of the learned embeddings for downstream ADL recognition. Such two-stage pipelines complicate deployment in real-world smart home systems. To overcome this limitation, we propose an end-to-end framework, \textbf{CARE (Contrastive Alignment for ADL Recognition)}, which integrates contrastive alignment and classification into a single-stage framework. In CARE, the model not only produces embeddings aligned across representation views but also outputs activity predictions directly, while classification learning simultaneously guides and regularizes representation learning. 

\noindent \textbf{Two-Branch Design.}  
CARE consists of two interconnected branches that share the same sequence and image encoders:  

\noindent -\emph{~Contrastive Representation Learning Branch}: This branch implements SICA. Encoders ($Enc_S(\cdot), Enc_I(\cdot)$) produce view-specific features ($\mathbf r_i^S, \mathbf r_i^I$), which are projected into a shared latent space via lightweight projections ($Proj_S(\cdot), Proj_I(\cdot)$). The SICA loss $\mathcal L_{SICA}$ enforces intra-view consistency and cross-view alignment. It is worth noting that both the sequence encoder and image encoder in CARE are modular components that can be flexibly replaced with alternative encoder architectures.

\noindent -\emph{~Classification Branch}: In parallel, CARE introduces a supervised 
classification head. The encoder outputs ($\mathbf r_i^S, \mathbf r_i^I$) are concatenated and passed to an MLP classifier $Cls_{SI}(\cdot)$ that predicts logits $\mathbf c$. The cross-entropy loss is defined as
\[
\mathcal L_{CE} = -\tfrac{1}{|B|}\sum\nolimits_{i=1}^{|B|} y_i \cdot 
\log(\sigma(Cls_{SI}(concat(\mathbf r_i^S,\mathbf r_i^I)))),
\]
where $\sigma$ denotes the softmax function and $y_i$ is the one-hot label for sample $i$. This branch provides task-level supervision and simultaneously regularizes the shared encoders, preventing them from overfitting to view-specific biases.  

\noindent \textbf{Joint Objective.}  
CARE jointly optimizes both branches:
\[
\label{eq:care}
\mathcal L_{CARE} = \beta \mathcal L_{SICA} + (1-\beta)\mathcal L_{CE},
\]
where $0\! \leq\!\beta\! \leq\! 1$ balances cross-view alignment and supervised 
classification. The classification head enables direct prediction and also provides gradients that complement SICA, improving representation discriminability in an end-to-end manner.  

Compared to SICA’s two-stage pipeline, CARE unifies representation alignment and 
classification into a single training objective. This design eliminates the 
representation–task mismatch, ensures that embeddings are both cross-view 
coherent and task-oriented, and yields a fully end-to-end system for robust ADL recognition.

\section{Experiments}

\begin{table*}[!hbtp]
  \centering
  \caption{Benchmarking on \textit{Milan}, \textit{Cairo}, and \textit{Kyoto7} (mean $\pm$ std over 5 runs). Time per batch measured on our setup.}
  \label{tab:overall}
  \resizebox{1.86\columnwidth}{!}{
  \setlength{\tabcolsep}{1mm}
\begin{threeparttable}
  \begin{tabular}{ccccccccccccc c}
    \toprule
    \multirow{2}{*}{Model} & \multicolumn{4}{c}{Milan} & \multicolumn{4}{c}{Cairo} & \multicolumn{4}{c}{Kyoto7} & \multirow{2}{*}{Time/Batch (s)}\\
    \cmidrule(l){2-5} \cmidrule(l){6-9} \cmidrule(l){10-13}
    & Acc. & Prec. & Rec. & F1 & Acc. & Prec. & Rec. & F1 & Acc. & Prec. & Rec. & F1 \\
    \midrule
    Random Forest & 32.0$\pm$9.6 & 40.0$\pm$4.4 & 37.8$\pm$6.2 & 34.8$\pm$4.9 & 61.7$\pm$9.8 & 53.9$\pm$9.4 & 57.4$\pm$5.5 & 52.6$\pm$5.4 & 43.2$\pm$4.5 & 46.8$\pm$5.6 & 44.6$\pm$3.8 & 41.4$\pm$4.3 & $\ll 0.001$\\
    SVM & 51.1$\pm$1.5 & 46.1$\pm$1.0 & 51.1$\pm$1.5 & 47.6$\pm$1.4 & 60.0$\pm$2.0 & 61.3$\pm$2.5 & 60.0$\pm$2.0 & 60.1$\pm$2.0 & 37.0$\pm$2.6 & 35.9$\pm$4.3 & 37.0$\pm$2.6 & 35.8$\pm$3.5 & $\ll 0.001$\\
    \midrule
    DeepCASAS~\cite{liciotti2020sequential} & 85.3$\pm$0.8 & 84.7$\pm$1.6 & 85.3$\pm$0.8 & 84.8$\pm$1.2 & 80.2$\pm$2.4 & 78.8$\pm$3.4 & 80.2$\pm$2.4 & 78.9$\pm$2.8 & 62.3$\pm$3.2 & 62.5$\pm$4.2 & 62.3$\pm$3.2 & 60.3$\pm$4.0 & 0.001\\
    JointTemporalModel\cite{app10155293} & 78.9$\pm$1.1 & 75.8$\pm$1.8 & 78.9$\pm$1.1 & 77.0$\pm$1.5 & 77.5$\pm$1.6 & 80.4$\pm$1.4 & 77.5$\pm$1.6 & 78.0$\pm$1.6 & 59.7$\pm$3.7 & 60.6$\pm$2.4 & 59.7$\pm$3.7 & 58.4$\pm$4.3 &0.001 \\
    EmbeddingFCN~\cite{bouchabou2021fully} & 72.8$\pm$1.6 & 72.0$\pm$1.5 & 73.0$\pm$1.4 & 71.0$\pm$1.6 & 48.3$\pm$3.5 & 37.1$\pm$3.3 & 48.1$\pm$3.4 & 40.0$\pm$3.4 & 59.3$\pm$2.0 & 57.1$\pm$2.0 & 59.0$\pm$2.1 & 55.5$\pm$2.2 & 0.005\\
    TCN~\cite{lea2016temporal} & 62.2$\pm$0.6 & 59.5$\pm$1.6 & 62.2$\pm$0.6 & 59.4$\pm$0.3 & 66.8$\pm$2.1 & 67.6$\pm$2.6 & 66.8$\pm$2.1 & 65.3$\pm$3.1 & 44.2$\pm$0.7 & 32.4$\pm$0.9 & 44.2$\pm$0.7 & 33.7$\pm$2.0 & 0.003\\
    GRU~\cite{DBLP:journals/corr/ChungGCB14} & 76.9$\pm$0.7 & 75.5$\pm$0.4 & 76.8$\pm$0.7 & 75.7$\pm$0.7 & \underline{83.0$\pm$0.0} & \underline{82.3$\pm$0.5} & \underline{82.7$\pm$0.5} & \underline{82.2$\pm$0.0} & 63.1$\pm$5.2 & 63.7$\pm$4.7 & 63.1$\pm$5.2 & 60.9$\pm$7.4 & 0.001\\
    1D-CNN~\cite{KIRANYAZ2021107398} & 46.2$\pm$2.2 & 44.0$\pm$2.8 & 46.2$\pm$2.2 & 44.5$\pm$2.4 & 56.7$\pm$3.3 & 57.7$\pm$2.5 & 56.7$\pm$3.3 & 56.8$\pm$3.2 & 42.7$\pm$3.4 & 40.4$\pm$5.1 & 42.7$\pm$3.4 & 40.6$\pm$4.3 & 0.001\\
    LSTM-CNN~\cite{9043535} & 71.2$\pm$5.7 & 67.6$\pm$5.8 & 71.2$\pm$5.7 & 69.0$\pm$5.8 & 62.3$\pm$1.1 & 56.9$\pm$3.8 & 62.3$\pm$1.1 & 58.5$\pm$1.0 & 50.8$\pm$2.1 & 51.5$\pm$1.9 & 50.8$\pm$2.1 & 45.2$\pm$4.5 & 0.003\\
    \midrule
    BinaryCNN~\cite{9245533} & 79.0$\pm$0.1 & 78.4$\pm$0.4 & 79.0$\pm$0.1 & 78.4$\pm$0.1 & 82.3$\pm$1.0 & 83.9$\pm$0.2 & 82.3$\pm$1.0 & 82.6$\pm$1.0 & 66.8$\pm$0.9 & 70.1$\pm$1.5 & 66.8$\pm$0.9 & 66.6$\pm$1.0 &0.003\\
    DCNN~\cite{gochoo2018unobtrusive} & 81.0$\pm$1.6 & 80.9$\pm$1.5 & 81.0$\pm$1.5 & 80.9$\pm$1.6 & 75.7$\pm$0.6 & 77.8$\pm$0.8 & 75.7$\pm$0.8 & 76.3$\pm$0.7 & 72.3$\pm$2.1 & 72.7$\pm$2.3 & 72.3$\pm$2.1 & 73.9$\pm$2.2 & 0.009\\
    GreyDCNN~\cite{mohmed2020employing} & 68.6$\pm$1.1 & 68.3$\pm$1.0 & 68.6$\pm$1.1 & 67.0$\pm$1.1 & 80.2$\pm$1.2 & 81.5$\pm$1.4 & 80.2$\pm$1.2 & 79.7$\pm$0.5 & 68.0$\pm$2.8 & 66.3$\pm$4.0 & 68.0$\pm$2.8 & 66.7$\pm$3.4 & 0.009\\
    Transformer~\cite{DBLP:journals/corr/abs-2010-11929} & 77.8$\pm$0.6 & 74.9$\pm$0.9 & 77.8$\pm$0.6 & 75.6$\pm$1.0 & 71.7$\pm$1.3 & 71.4$\pm$1.5 & 71.6$\pm$1.6 & 70.8$\pm$1.8 & 56.1$\pm$0.0 & 58.7$\pm$4.4 & 56.1$\pm$0.0 & 56.1$\pm$2.3 & 0.007\\
    \midrule
    GraphModel~\cite{s24123944} & 82.8$\pm$0.5 & 82.3$\pm$0.2 & 82.8$\pm$0.5 & 82.3$\pm$0.3 & 76.7$\pm$0.5 & 77.1$\pm$0.3 & 76.7$\pm$0.2 & 76.8$\pm$0.5 & 53.4$\pm$1.2 & 54.2$\pm$1.0 & 53.4$\pm$1.0 & 52.0$\pm$1.2 & 1.586\\
    \midrule
    TDOST~\cite{TDOST} & \underline{88.7$\pm$1.0} & \underline{88.5$\pm$0.9} & \underline{88.7$\pm$1.0} &\underline{88.4$\pm$0.9}   & 81.0$\pm$0.9 & 80.3$\pm$1.9 & 81.0$\pm$0.9 & 79.3$\pm$1.9 & 70.0$\pm$1.2 & 70.0$\pm$1.1 & 70.0$\pm$1.2 & 70.0$\pm$1.1 & 0.001\\
    {CPC~\cite{haresamudram2021contrastive}} & 85.9$\pm$0.1 & 86.3$\pm$0.8 & 86.0$\pm$0.5 & 85.7$\pm$0.7 & 73.7$\pm$0.5 & 73.0$\pm$0.7 & 73.7$\pm$0.5 & 72.5$\pm$0.2 & \textbf{73.2$\pm$4.3} & \textbf{73.2$\pm$5.2} &
\textbf{73.1$\pm$4.3} & \textbf{73.0$\pm$4.4}
 & $\ll 0.001$\\
    \midrule
    \textbf{SICA}\textsubscript{L-Res} & \cellcolor{lightgreen}\textbf{89.8$\pm$0.3} & \cellcolor{lightgreen}\textbf{89.8$\pm$0.2} & \cellcolor{lightgreen}\textbf{89.8$\pm$0.3} & \cellcolor{lightgreen}\textbf{89.6$\pm$0.2} & \textbf{88.5$\pm$0.7} & \textbf{89.3$\pm$0.5} &
\textbf{88.5$\pm$0.7} & \textbf{88.6$\pm$0.6}
& \underline{72.6$\pm$1.8} & \underline{73.9$\pm$2.4} & \underline{72.6$\pm$1.8} & \underline{72.4$\pm$1.7} & 0.010\\
   \textbf{SICA}\textsubscript{BiL-Res} &
\textbf{89.5$\pm$0.6} &
\textbf{89.7$\pm$0.5} &
\textbf{89.5$\pm$0.6} &
\textbf{89.5$\pm$0.5} &
\cellcolor{lightgreen}\textbf{88.9$\pm$1.3} &
\cellcolor{lightgreen}\textbf{88.8$\pm$1.9} &
\cellcolor{lightgreen}\textbf{88.9$\pm$1.3} &
\cellcolor{lightgreen}\textbf{88.7$\pm$1.6} &
\cellcolor{lightgreen}\textbf{73.3$\pm$2.6} &
\cellcolor{lightgreen}\textbf{74.9$\pm$1.7} &
\cellcolor{lightgreen}\textbf{73.8$\pm$2.5} &
\cellcolor{lightgreen}\textbf{73.3$\pm$2.6} & 0.011 \\
    \bottomrule
\end{tabular}
\end{threeparttable}
}
\end{table*}

\subsection{Datasets and Experimental Protocol}
\textbf{Datasets.} We evaluate on three CASAS datasets—\textit{Milan}, \textit{Cairo}, and \textit{Kyoto7}—collected in realistic smart-home testbeds with bedrooms, bathrooms, kitchens, living rooms, and dining rooms instrumented by ambient sensors (motion, temperature, door), following Cook \emph{et al.}~\cite{cook2012casas}. 
\textit{Milan} contains 35 sensors, 15 activity categories, and 4{,}060 recorded activities; \textit{Cairo} has 34 sensors, 13 categories, and 842 activities; \textit{Kyoto7} includes 71 sensors, 13 categories, and 634 activities. For comparability, we group the original labels into {10 ADLs with an additional null (“Other”) class, following the standard protocol in prior work (e.g.,~\cite{liciotti2020sequential,TDOST,gochoo2018unobtrusive,mohmed2020employing,s24123944}).}

\textbf{Data splits.} We adopt stratified 70/30 train/test split by activity class. On training portion, we run 5-fold cross-validation for hyperparameter tuning, then retrain on the full training split and evaluate on the held-out test split. Each experiment is repeated with 5 random seeds; we report mean~$\pm$~std. 

\textbf{Batching and training.} Batch size is 64; training runs up to 60 epochs with early stopping on validation F1. No random window sampling is used. We use a learning rate of 0.001, weight decay of 0.001, and no learning-rate scheduling. The temperature of the supervised contrastive loss is set to 0.05.

{ \textbf{Model structure.} Our sequence/image encoders (Sec.~\ref{sec:seq}, Sec.~\ref{sec:img}) are \textbf{architecture-agnostic} that can be instantiated with any sequence or image backbone. In experiments, we use \emph{LSTM} or \emph{BiLSTM} for the sequence stream and \emph{ResNet18} for the image stream; the corresponding dual-stream variants are denoted \textbf{L-Res} and \textbf{BiL-Res}. Both streams are projected to a 32-d shared embedding for contrastive alignment, and an MLP classifier predicts labels from the concatenated features. 
Implementation details are in the released codebase.}

\subsection{Overall Performance}
\label{sec:overall}

\noindent \textbf{Baselines.} We compare our method against five categories of baselines:  
(1) \emph{Traditional ML:} Random Forest, Support Vector Machines (SVM);  
(2) \emph{Sequence-based:} DeepCASAS~\cite{liciotti2020sequential}, Embedding FCN~\cite{bouchabou2021fully}, TCN~\cite{lea2016temporal}, GRU~\cite{DBLP:journals/corr/ChungGCB14}, 1D-CNN~\cite{KIRANYAZ2021107398}, LSTM-CNN~\cite{9043535};  
(3) \emph{Image-based:} DCNN~\cite{gochoo2018unobtrusive}, GreyDCNN~\cite{mohmed2020employing}, Transformer~\cite{DBLP:journals/corr/abs-2010-11929};  
(4) \emph{Graph-based:} GraphModel~\cite{s24123944}; {(5)  \emph{Contrastive-based:} CPC~\cite{haresamudram2021contrastive}};
(6) \emph{Language-inspired:} TDOST~\cite{TDOST}, which frames activity recognition under a language modeling paradigm.  This diverse set of baselines covers traditional feature-based classifiers, sequence/image/graph encoders, and recent language-inspired models, providing a comprehensive benchmark for evaluating our framework.
All baseline and proposed models are trained and evaluated under the \emph{exact same} train/test splits and evaluation protocol, ensuring fairness and eliminating confounding factors due to differing data partitions.

\noindent \textbf{Overall performance.} Table~\ref{tab:overall} compares all approaches on \textit{Milan}, \textit{Cairo}, and \textit{Kyoto7}. Traditional machine learning models, such as Random Forest and SVM, perform significantly worse than deep learning baselines, underscoring the necessity of representation learning in complex ADL dynamics. 
 Deep learning approaches yield substantially higher performance. Sequence-based baselines (e.g., DeepCASAS, EmbeddingFCN, GRU, LSTM-CNN) achieve competitive results as they usually capture temporal dependencies effectively, but are sometimes more affected by noise and sparsity. While image-based baselines (e.g., BinaryCNN, DCNN, GreyDCNN, Transformer) are generally more robust but may degrade on highly sparse datasets such as \textit{Kyoto7}. This complementarity highlights the importance of integrating both views. The graph-based model performs competitively on \textit{Milan} and \textit{Cairo} by explicitly modeling sensor relationships, but suffers a marked performance degradation on the sparse and noisy \textit{Kyoto7}. By using a textual encoding method, the accuracy of TDOST is generally better than that of all the other baselines except ours, demonstrating the advantage of textual information in capturing cross-view information. {The contrastive-based method, CPC, shows competitive performance on \textit{Kyoto7}, but falls short on \textit{Milan} and \textit{Cairo} compared with our method and TDOST.}
Overall, our SICA models consistently deliver superior performance across all datasets and metrics, establishing a new state of the art.

\noindent \textbf{Computation considerations.} 
In general, image-based models require more computation than sequence-based ones due to the inherent differences in input size and model complexity. GraphModel further exacerbates this with prohibitively high inference time (1.586 s/batch). TDOST consumes only 0.001 seconds per batch by using BiLSTM for inference. 
Despite leveraging both representation views, our SICA models remain efficient (0.010–0.011 s/batch), on par with image-based baselines and an order of magnitude faster than GraphModel. 
This favorable trade-off between accuracy and efficiency underscores SICA's scalability for real-world ADL recognition.

\subsection{Uni- vs. Within- and Cross-View Alignment}

In our supervised contrastive learning framework, we consider two types of positive pairs: (1) \textit{intra-view} pairs, i.e., embeddings from the same view and class (sequence–sequence or image–image); and (2) \textit{cross-view} pairs, i.e., embeddings across views but from the same class. The latter includes both the direct sequence–image pair of the same sample ($z_i^S, z_i^I$) and cross-sample pairs from the same class ($z_i^S, z_j^I$ with $y_j=y_i$).

\begin{table}[htbp]
  \centering
  \caption{Ablation: uni-, within-, and cross-view alignments: performance metrics (mean $\pm$ std)) and computational cost in memory (MiB) and time per batch (s).}
  \resizebox{1\columnwidth}{!}{
  \setlength{\tabcolsep}{1mm}
  \begin{tabular}{c|c|ccccccccc}
    \toprule
    & \multirow{2}{*}{Metric} & \multicolumn{3}{c}{Uni-View} & \multicolumn{2}{c}{Within-View Align} &  \multicolumn{2}{c}{Self-Supervised Align} & \multicolumn{2}{c}{Cross-View Align}\\
    \cmidrule(l){3-5}\cmidrule(l){6-7}\cmidrule(l){8-9}\cmidrule(l){10-11}
    & & LSTM & BiLSTM & ResNet18 & L-Res & BiL-Res& L-Res & BiL-Res & L-Res & BiL-Res \\
    \midrule
    \multirow{4}{*}{\rotatebox{90}{Milan}}
      & Acc.   & 86.3$\pm$0.8 & 86.3$\pm$2.1 & 87.0$\pm$0.8 & 89.5$\pm$0.3 & 89.6$\pm$0.7 & 
      \textbf{91.3$\pm$0.7}
      & \textbf{92.9$\pm$0.3} & {89.8$\pm$0.3}& {89.5$\pm$0.6} \\
      & Prec.  & 86.0$\pm$0.9 & 86.1$\pm$2.2 & 86.8$\pm$1.0 & 89.6$\pm$0.4 & 89.9$\pm$0.6 & 
      \textbf{{91.2$\pm$0.8}}
      & \textbf{92.6$\pm$0.8} &{89.8$\pm$0.2 }& {89.7$\pm$0.5} \\
      & Rec.   & 86.2$\pm$0.8 & 86.3$\pm$2.2 & 87.0$\pm$0.8 & 89.5$\pm$0.3 & 89.6$\pm$0.7 & 
     \textbf{91.3$\pm$0.7}
      & \textbf{92.9$\pm$0.3} & {89.8$\pm$0.3} & {89.5$\pm$0.6} \\
      & F1     & 85.9$\pm$0.9 & 86.1$\pm$2.2 & 86.4$\pm$1.0 & {89.3$\pm$0.3} & 89.5$\pm$0.6 &
      \textbf{90.8$\pm$0.8}
      & \textbf{92.5$\pm$0.5} &{89.6$\pm$0.2} & {89.5$\pm$0.5} \\
    \midrule
    \multirow{4}{*}{\rotatebox{90}{Cairo}}
      & Acc.   & 83.4$\pm$4.1 & 83.8$\pm$4.3 & 76.7$\pm$3.8 & 88.1$\pm$0.4 & 88.1$\pm$0.4 & 
      81.2$\pm$0.3
      & 86.6$\pm$1.0 &\textbf{88.5$\pm$0.7} & \textbf{88.9$\pm$1.3} \\
      & Prec.  & 84.5$\pm$3.1 & 85.2$\pm$3.8 & 81.6$\pm$1.6 & 87.6$\pm$0.2 & 88.6$\pm$1.3 & 
      80.8$\pm$0.7
      & 87.2$\pm$1.2 &\textbf{89.3$\pm$0.5} & \textbf{88.8$\pm$1.9} \\
      & Rec.   & 83.9$\pm$4.5 & 84.5$\pm$4.4 & 76.7$\pm$3.8 & 88.1$\pm$0.4 & 88.1$\pm$0.4 &
      81.2$\pm$0.3
      & 86.6$\pm$1.0 &\textbf{88.5$\pm$0.7} & \textbf{88.9$\pm$1.3} \\
      & F1     & 82.3$\pm$6.1 & 84.1$\pm$4.5 & 77.4$\pm$3.5 & 87.8$\pm$0.2 & 88.1$\pm$0.7 &
      80.0$\pm$0.3
      & 86.3$\pm$1.2 & \textbf{88.6$\pm$0.6} & \textbf{88.7$\pm$1.6} \\
    \midrule
    \multirow{4}{*}{\rotatebox{90}{Kyoto7}}
      & Acc.   & 71.4$\pm$4.0 & 72.4$\pm$2.7 & 71.7$\pm$2.7 & 72.5$\pm$1.3 & 72.0$\pm$2.9 & 
      72.0$\pm$0.9
      & 72.5$\pm$1.3 & \textbf{72.6$\pm$1.8} & \textbf{73.3$\pm$2.6} \\
      & Prec.  & 72.0$\pm$6.0 & 72.2$\pm$2.2 & 70.6$\pm$3.2 & 72.9$\pm$1.6 & 73.5$\pm$2.3 &
      73.1$\pm$1.2
      & {74.3$\pm$2.4} & \textbf{73.9$\pm$2.4} & \textbf{74.9$\pm$1.7} \\
      & Rec.   & 72.0$\pm$4.0 & 72.2$\pm$2.2 & 71.7$\pm$2.7 & 72.5$\pm$1.3 & 72.0$\pm$3.0 &
      72.0$\pm$0.9
      & 72.5$\pm$1.3 & \textbf{72.6$\pm$1.8} & \textbf{73.8$\pm$2.5} \\
      & F1     & 71.0$\pm$4.0 & 71.1$\pm$2.2 & 71.1$\pm$3.7 & 71.8$\pm$1.5 & 72.0$\pm$3.2 & 
      71.2$\pm$1.3
      & 71.8$\pm$1.8 & \textbf{72.4$\pm$1.7} & \textbf{73.3$\pm$2.6} \\
    \midrule
    \multirow{2}{*}{\rotatebox{90}{Eff.}}
      & Mem.     & 171.76 & 313.53 & 1511.19 & 1620.17 & 1716.93   &1620.17 & 1716.93& 1620.17 & 1716.93 \\
      & T/B  & 0.004  & 0.005  & 0.006   & 0.010   & 0.011   & 0.010   & 0.011 & 0.010   & 0.011   \\
    \bottomrule
  \end{tabular}}
  \label{tab:single_align}
\end{table}


We evaluate four variants: (i) \textbf{uni-view baselines} (LSTM, BiLSTM, ResNet18), which learn within a single domain; (ii) \textbf{within-view alignment}, which applies contrastive loss only to intra-view pairs; (iii) \textbf{cross-view alignment} (our full SICA), which leverages both intra- and cross-view pairs; and {(iv) \textbf{self-supervised alignment} can be regarded as an \textit{instance-level cross-view alignment}, which aligns the two views using an instance-level contrastive objective: only the paired embeddings of the same sample are treated as positive pairs, without using class labels to construct cross-sample positives. }

As shown in Tab.~\ref{tab:single_align}, cross-view alignment consistently outperforms all baselines. For instance, BiL-Res achieves gains of $\sim$3\% on \textit{Milan}, 2\% on \textit{Cairo}, and 1\% on \textit{Kyoto7} compared with the best unimodal models.
{Compared with within-view alignment, cross-view alignment yields comparable or up to $\sim1\%$ higher accuracy, and is our default due to its consistent gain across datasets. Instance-level cross-view alignment is competitive on \textit{Milan} but degrades on smaller datasets. Both validate the effectiveness of our cross-view design.}

 \begin{figure}[htbp]
    \centering
    \vspace{-1mm}
    \begin{subfigure}[b]{0.45\textwidth}
\includegraphics[width=\textwidth]{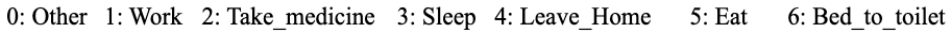}
        \label{fig:bar}
        \vspace{-3mm}
    \end{subfigure}
    \begin{subfigure}[b]{0.158\textwidth}
        \includegraphics[width=\textwidth]{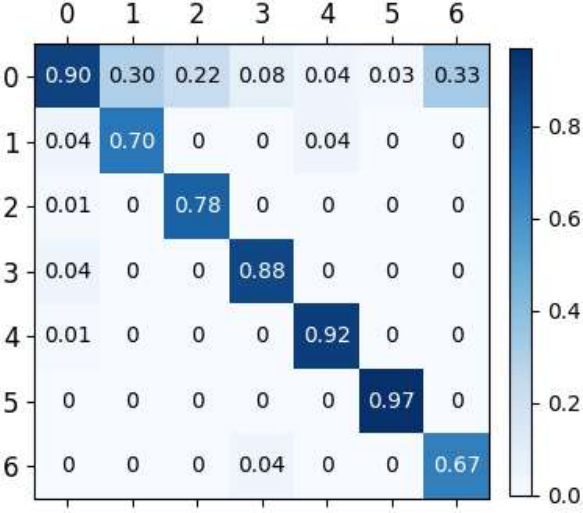}
        \caption{BiL-Res}
        \label{fig:BR}
    \end{subfigure}
    \begin{subfigure}[b]{0.158\textwidth}
        \includegraphics[width=\textwidth]{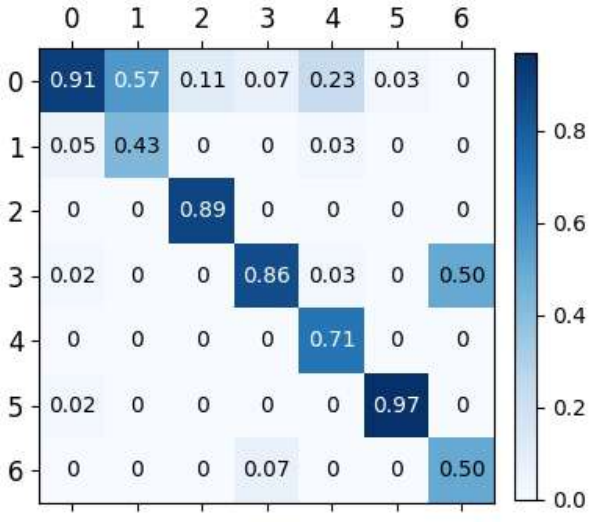}
        \caption{BiLSTM}
        \label{fig:B}
    \end{subfigure}
    \begin{subfigure}[b]{0.158\textwidth}
        \includegraphics[width=\textwidth]{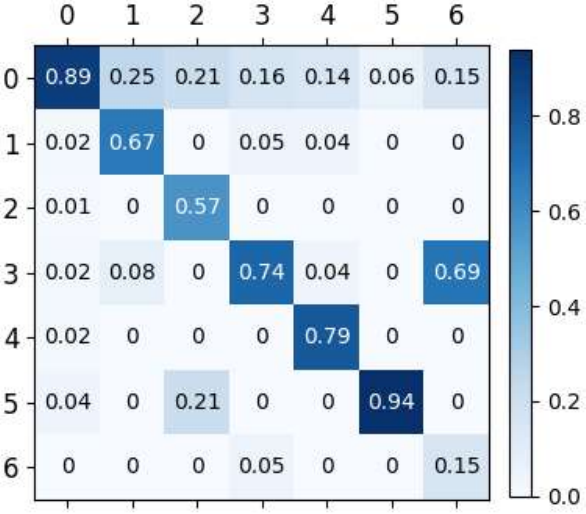}
        \caption{ResNet18}
        \label{fig:R}
    \end{subfigure}
    
    \caption{Heatmap of BiLSTM and ResNet18 on \textit{Cairo}.}
    \label{fig:cairo_heatmap}
    \end{figure}    

Fig.~\ref{fig:cairo_heatmap} presents \textit{Cairo} as a representative example, providing class-level evidence of our approach. The results show that SICA leverages the complementary strengths of sequence and image streams: across all activities, it either outperforms both unimodal baselines or matches the stronger one, achieving clear gains on “Work,” “Sleep,” “Leave Home,” and "Bed to toilet," while remaining competitive on classes such as “Take medicine” and “Eat.” This consistent per-class advantage highlights how cross-view alignment fully exploits complementary temporal and spatial cues, underscoring the robustness of our cross-view approach.

In terms of efficiency, multimodal models require more memory (e.g., $\sim$1.6 GB for ResNet18/SICA vs. 0.17 GB for LSTM), yet their runtime overhead remains modest (0.010–0.011 s/batch), effectively matching ResNet18. This makes the performance–efficiency tradeoff favorable.

\subsection{Contrastive Alignment vs. Naïve Fusion}

A natural baseline for multimodal integration is to concatenate the sequence and image embeddings and train with a cross-entropy (CE) loss. While straightforward, this naïve fusion provides no explicit signal to align the two views, leaving the classifier free to overfit to the dominant stream and underutilize complementary cues. 

\begin{figure}[tb]
  \centering
  \includegraphics[width=1.0\linewidth]{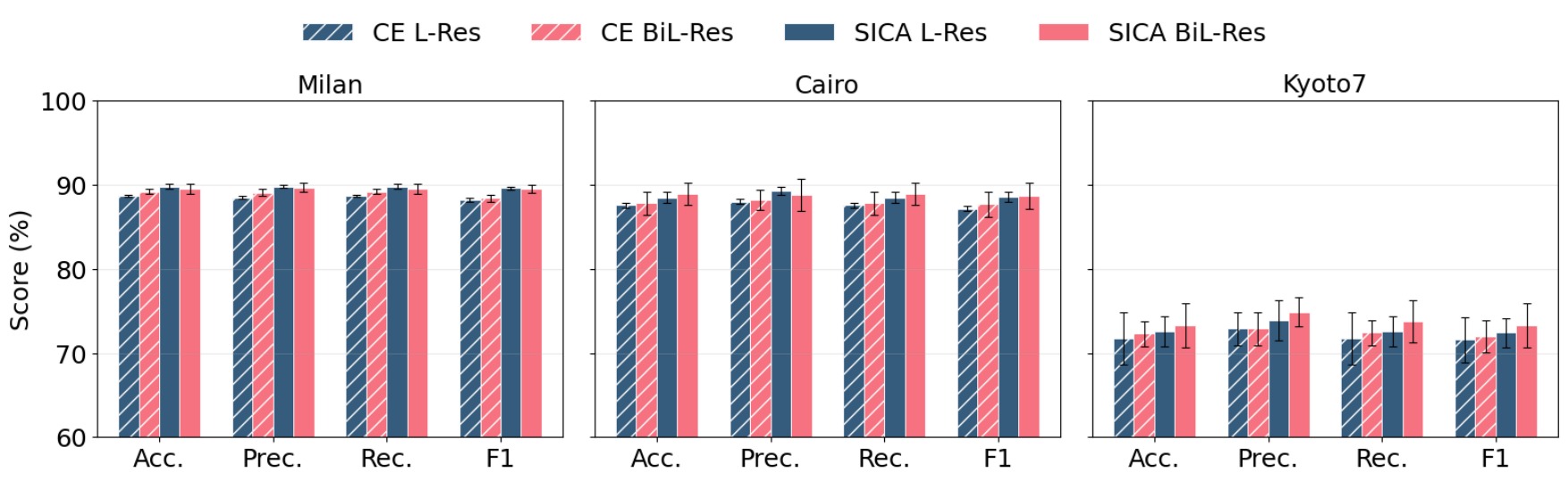}
  \caption{Ablation: naïve fusion via cross entropy (CE) vs. contrastive alignment (SICA).}
  \label{fig:cevssica}
\end{figure}

Fig.~\ref{fig:cevssica} compares the two approaches across three datasets. SICA has a comparable performance with CE-based fusion, with the largest gains observed on the challenging \textit{Kyoto7} dataset. For example, BiL-Res with SICA improves the F1-score by 1.3\% over its CE counterpart. Importantly, these CE models are not previous baselines but architectural counterparts of our framework, differing only in the training objective—underscoring the benefit of contrastive alignment.

The performance gap is most pronounced in \textit{Kyoto7}, which combines limited activity samples ($\sim$600 instances) with a dense set of 71 sensors, resulting in higher dimension, sparser and noisier inputs compared with the other two datasets. Such conditions hinder naïve concatenation, as the classifier struggles to disentangle redundant activations and overfits view-specific cues. SICA alleviates this by aligning sequence and image representations in a shared latent space, leveraging spatial images to filter redundant activations and enforcing view-invariant, class-consistent features.

\begin{figure}[tb]
  \centering
  \includegraphics[width=1.0\linewidth]{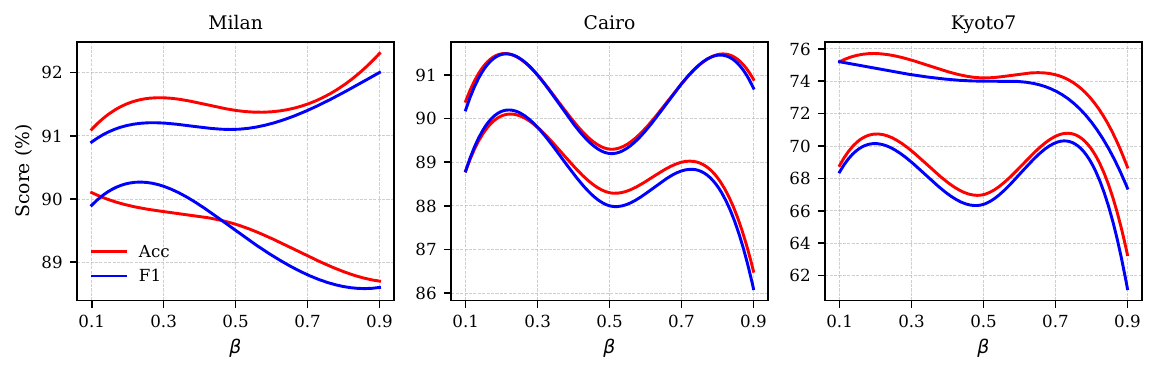}
    \caption{Sensitivity Analysis of $\beta$ for BiL-Res.}
  \label{fig:sen}
\end{figure}

{ Furthermore, in Eq.~\ref{eq:care}, the hyperparameter $\beta$ controls the extent to which SICA deviates from naive fusion. We evaluate BiL-Res under different $\beta$ and report the corresponding mean $\pm$ std bounds in Fig.~\ref{fig:sen}. Across all datasets, performance varies within approximately 1\% on \textit{Milan}, 2\% on \textit{Cairo}, and 4\% on \textit{Kyoto7}, demonstrating that CARE is relatively insensitive to the choice of $\beta$ and robust to moderate variations of it.}

\noindent \textbf{Efficiency.} Both CE fusion and SICA share the same input and dual-stream encoders, and differ only in the learning objective. As a result, their computational cost is identical, ensuring that SICA’s performance advantage stems from improved representation learning rather than architectural complexity.

Overall, these results confirm that contrastive alignment is not merely an alternative to feature concatenation but a principled mechanism for exploiting cross-view complementarities. By explicitly aligning sequence and image embeddings, SICA yields more robust and generalizable representations, particularly under the data sparsity and noise conditions common in real-world smart home deployments.

\subsection{Temporal Binning and Frequency-based Event Filtering}
\label{sec:data}

\begin{figure}[tb]
  \centering
  \includegraphics[width=1.0\linewidth]{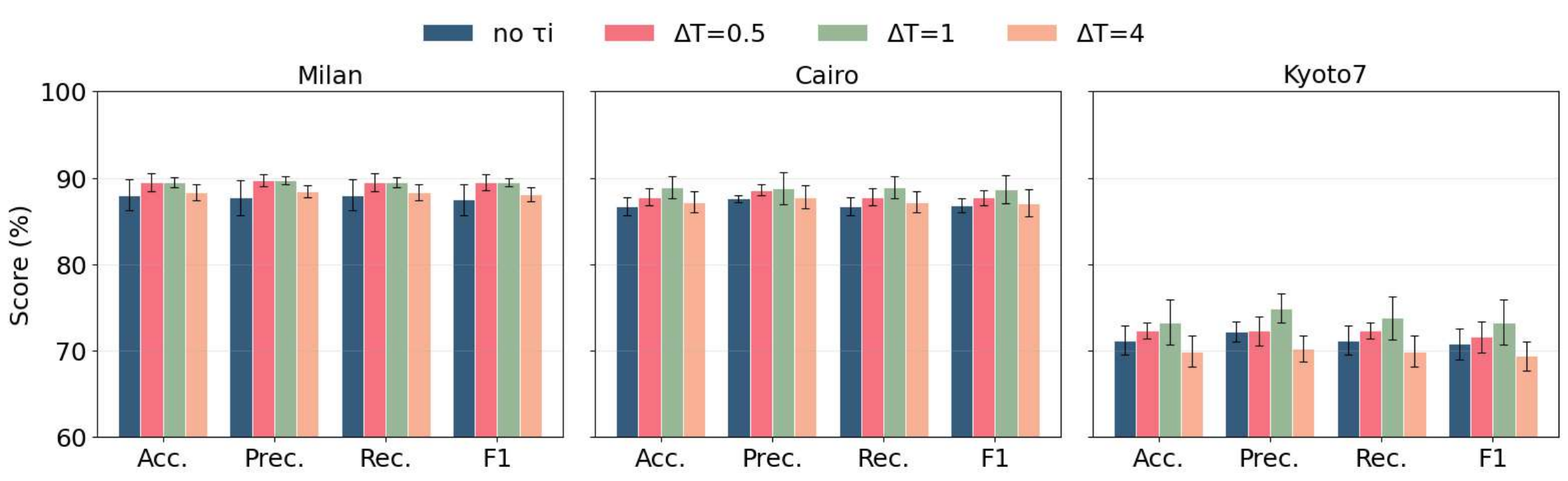}
  \caption{Performance Comparison Using Different Hour Intervals for Temporal Encoding of BiL-Res.}
  \label{fig:timehour}
\end{figure}

\noindent \textbf{Temporal binning.}
We first examine the impact of discretizing timestamps into coarse-grained bins. Fig.~\ref{fig:timehour} reports results with bin widths of 0.5h, 1h, and 4h ($\Delta T\in\{0.5,1,4\}$), compared to the case without incorporating time information. The results without incorporating time information is lower than all the other settings by 1--3\%, suggesting the importance of timestamps. A 1-hour bin consistently yields the best trade-off across all datasets: finer bins (0.5h) introduce redundancy and unstable variance (e.g., on \textit{Kyoto7}), while coarser bins (4h) discard critical context between adjacent activities. This aligns with the average activity durations (0.25h for \textit{Milan}, 1.09h for \textit{Cairo}, and 0.67h for \textit{Kyoto7}), confirming the 1-hour setting best balances temporal resolution and robustness. {We also find that without temporal binning, the model remains competitive, suggesting binning is not the sole driver of performance.}

\begin{figure}[tb]
    \centering
    \vspace{-1mm}
    \begin{subfigure}[b]{0.2\textwidth}
\includegraphics[width=0.9\textwidth]{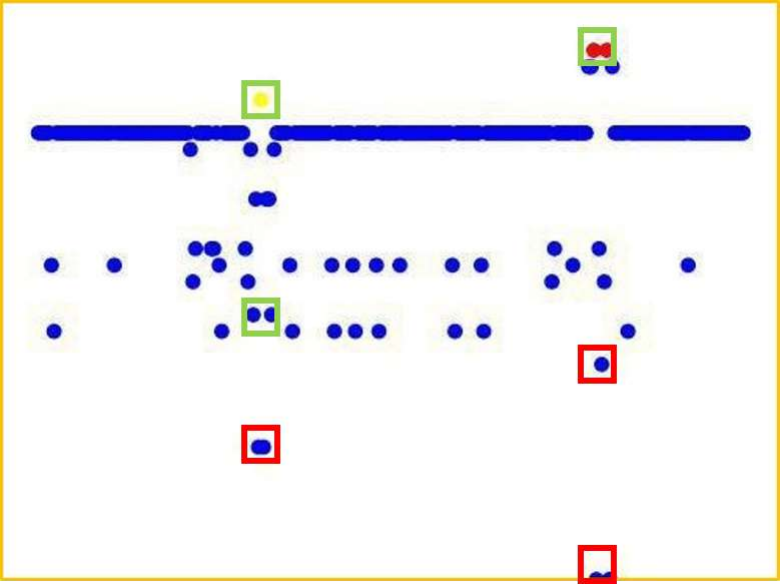}
        \caption{w/o FEF}
        \label{fig:ex1}
    \end{subfigure}
    \begin{subfigure}[b]{0.2\textwidth}
        \includegraphics[width=0.9\textwidth]{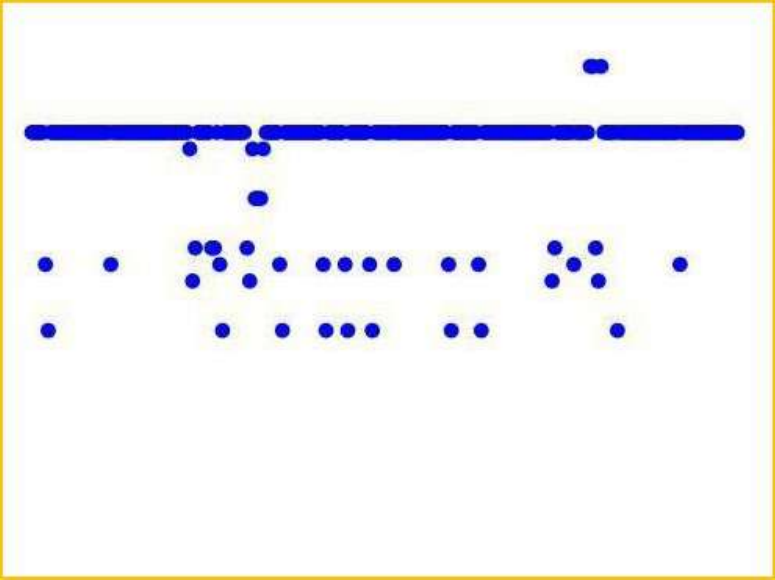}
        \caption{w FEF}
        \label{fig:ex2}
    \end{subfigure}
    \caption{Event Filtering Example: Eat activity in \textit{Milan}.}
    \label{fig:cairo_filter}
\end{figure}

\noindent \textbf{Frequency-based Event Filtering (FEF).}
Figure~\ref{fig:cairo_filter} illustrates the effect of FEF on the \textit{Milan} dataset. The filter discards several sensors—shown in red, yellow and some blue—including "M009", "M006" and "M019" in red box (located in the living room and aisle), as well as "M012", "D003" and "T001" in green box (placed in the kitchen). Although FEF eliminates sensors located in the kitchen, which might seem relevant to the "Eat" activity, the triggered frequency of these sensors accounts for less than 7\% of the total sensor activations during "Eat". Besides, the accuracy for "Eat" increases from 66.7\% to 80.6\% by using FEF, indicating that the sensors in the green box are not the most important features for this activity.

Consequently, FEF eliminates sensors that are highly irrelevant to the target activity; however, it may also remove useful information. Therefore, an appropriate filtering threshold must be carefully chosen.
We also test the effect of removing sensors with very low activation frequency. 
As shown in Table~\ref{tab:frequency}, the best performance on \textit{Cairo} is achieved with a 1\% threshold; results obtained without FEF or at higher thresholds exhibit a slight decline. This suggests that filtering suppresses spurious noise while preserving salient sensor events, with 1\% threshold providing the best robustness–information balance. 
{To further isolate FEF's contribution, we integrate it into DeepCASAS. This boosts accuracy by $\sim$3\% on \textit{Milan} and $\sim$8\% on \textit{Kyoto7}, supporting FEF’s effectiveness and generality. In contrast, the gain on \textit{Cairo} is marginal ($<$1\%), suggesting preprocessing alone cannot account for the improvement. Collectively, the gains are driven by CARE’s overall design.}


\begin{table}[tb]
  \centering
  \caption{BiL-Res on \textit{Cairo}: FEF Threshold $\theta$.}
  \resizebox{0.6\columnwidth}{!}{
  \setlength{\tabcolsep}{1mm}
  \begin{tabular}{c cccc}
    \toprule
    $\theta$  & Acc. & Prec. & Rec. & F1 \\
    \midrule
    W/O FEF & 87.0$\pm$2.4 & 87.2$\pm$2.7 & 87.0$\pm$2.4 & 86.7$\pm$2.4\\ 
    1\%  & \textbf{88.9$\pm$1.3} & \textbf{88.8$\pm$1.9} & \textbf{88.9$\pm$1.3} & \textbf{88.7$\pm$1.6} \\
    3\% & 87.0$\pm$1.8 & 87.3$\pm$1.6 & 87.0$\pm$1.8 & 86.6$\pm$1.9\\
    5\%  &  87.6$\pm$0.5 & 88.1$\pm$0.3 & 87.6$\pm$0.5 & 87.3$\pm$0.7\\
    \bottomrule
  \end{tabular}}
  \label{tab:frequency}
\end{table}

{We acknowledge that the temporal bin width and frequency thresholds are manually selected, which may limit the generality of the method. However, ablations show the method is not sensitive to their choices (accuracy varies by $<3\%$ across datasets), so coarse tuning is sufficient in practice). } Together, the ablations confirm that our preprocessing pipeline—hourly temporal binning plus Frequency-based Event Filtering—strikes the right compromise between noise suppression and information retention, providing a clean foundation for downstream multimodal representation learning.

\subsection{Effectiveness of Image-Based Representations}

\begin{figure}[tb]
  \centering
  \includegraphics[width=1.0\linewidth]{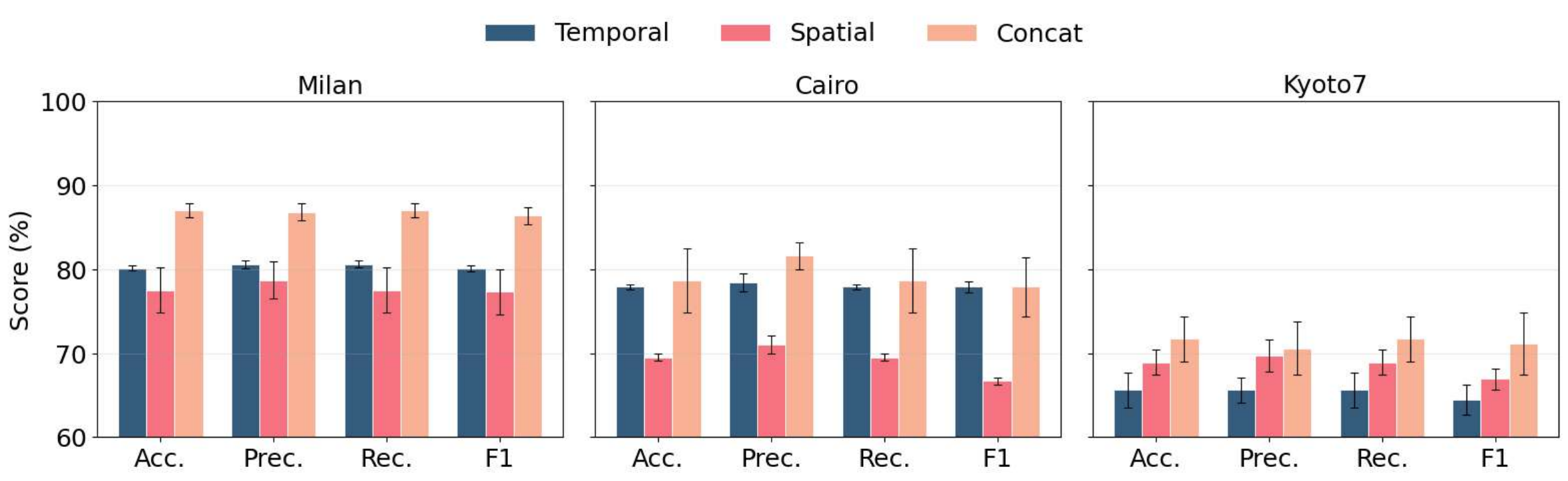}
  \caption{ResNet18 Ablation on Image Inputs: temporal vs. spatial vs. concatenated. }
  \label{fig:image_abl}
\end{figure}

To interpret the contributions of temporal and spatial cues in image encodings, we train ResNet18 on temporal images, spatial images, and their concatenation. Fig.~\ref{fig:image_abl} report the results. Concatenated images consistently outperform single-view inputs, with 5--7\% gains on \textit{Milan} and 1--8\% on \textit{Cairo}. These results highlight the complementary nature of the two views: temporal images capture fine-grained activity dynamics, while spatial images provide global context via sensor layout and transition statistics. On \textit{Kyoto7}, the gain from concatenation is smaller. This is likely due to its denser sensor deployment and higher noise levels, which make the marginal contribution of explicit spatial images less pronounced. Nevertheless, concatenation never underperforms compared to single views, confirming that temporal--spatial image 
integration yields more robust and generalizable representations. Crucially, the gains stem from the representation design itself rather than the backbone architecture, underscoring the effectiveness and broad applicability of our image encoding strategy.

    \begin{figure}[tb]
    \centering
    \begin{subfigure}[t]{0.14\textwidth}
        \centering
        \includegraphics[width=\linewidth]{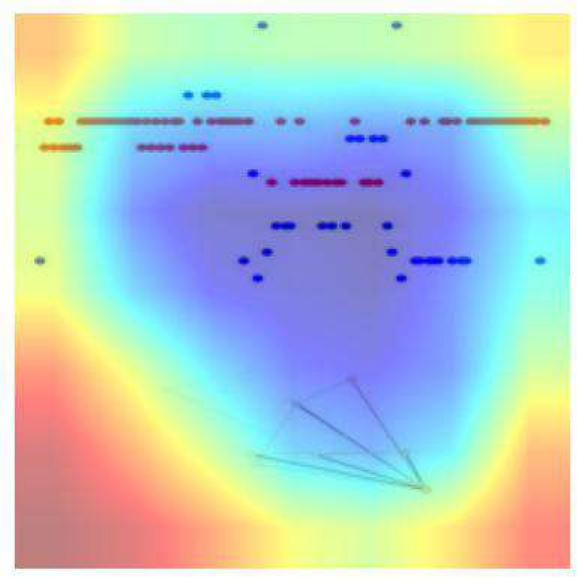}
        \label{fig:milan_cam1}
    \end{subfigure}
    \hfill
    \begin{subfigure}[t]{0.14\textwidth}
        \centering
        \includegraphics[width=\linewidth]{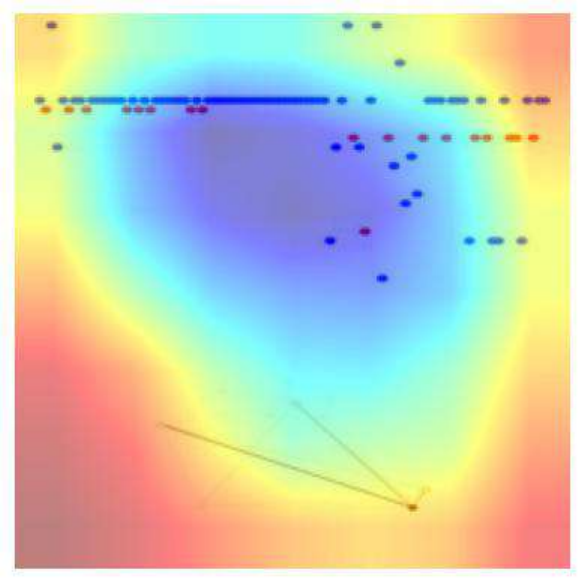}
        \label{fig:milan_cam2}
    \end{subfigure}
    \hfill
    \begin{subfigure}[t]{0.14\textwidth}
        \centering
        \includegraphics[width=\linewidth]{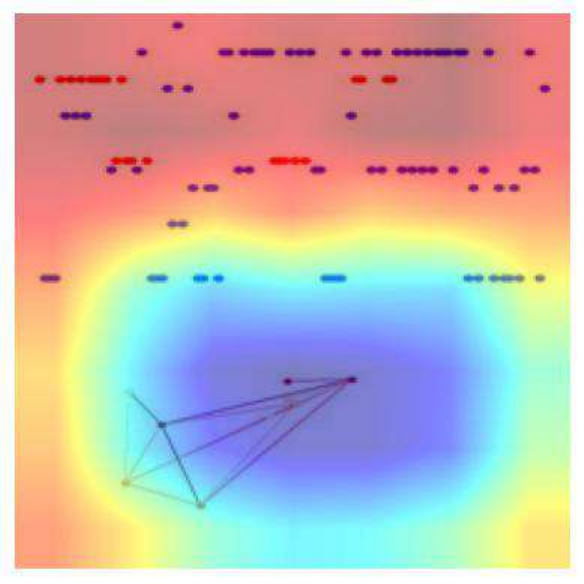}
        \label{fig:milan_cam3}
    \end{subfigure}
    \vspace{-4mm}
    \caption{\textit{Cairo}'s Grad-Cam on Work Activity.}\label{fig:milan_cam}
    \end{figure}

To understand the role of different cues, we employ Gradient-weighted Class Activation Mapping (Grad-CAM)~\cite{selvaraju2017grad}, a visualization technique that leverages the gradients of a target class flowing into a convolutional layer to generate a heatmap of salient regions for prediction. As shown in Fig.~\ref{fig:milan_cam}, the heatmaps (temporal image on the top and spatial image on the bottom) indicate that temporal images highlight sequential activation patterns, while spatial images emphasize sensor layout regions relevant to transitions. The model distributes its attention across either/both temporal or/and spatial areas, demonstrating that concatenation encourages the network to leverage complementary information sources. This visualization corroborates the quantitative results and explains the consistent performance gains of the joint representation.

\subsection{Robustness to Sensor Corruption, Layout Variability and Temporal Drift}

{Real-world deployments face distribution shifts from sensor malfunctions, missing signals, sensor layout changes, and temporal drift. We train on clean data and evaluate directly on corrupted test sets \emph{\underline{without retraining}} to test robustness to corruption and layout variability. For temporal drift, we adopt a time-aware split: train on the earlier $70\%$ and test on the remaining $30\%$.}

\noindent \textbf{Experimental setup.} We consider two common types of corruption: (i) \textit{Sensor malfunctions}, where a fraction of activity samples are randomly selected and a percentage of their events are replaced with default values (``OFF," ``CLOSE," ``ABSENT") to simulate faulty or missing sensors. (ii) \textit{Sensor repositioning}, where sensor coordinates in spatial images are perturbed by Gaussian noise with variance $\sigma^2\!\in\!\{10,20,30\}$, mimicking layout changes or misalignments. 

\begin{table}[tb]
  \centering
  \caption{Performance of BiL-Res under sensor corruption and positional perturbations. 
For corruption settings $a\%\!\times\!b\%$, $a\%$ of activity samples are selected and $b\%$ of their sensor events replaced with default values.}
  \resizebox{1\columnwidth}{!}{
  \setlength{\tabcolsep}{1mm}{
  \begin{tabular}{c|c|ccccccc}
    \toprule
    \multirow{2}{*}{} &\multirow{2}{*}{Metric}  & \multirow{2}{*}{None} & \multicolumn{3}{c}{Malfunctions and Varying Numbers} & \multicolumn{3}{c}{Sensor Repositioning}\\
    \cmidrule(l){4-6} \cmidrule(l){7-9}
    &&&$5\%\times5\%$ & $5\%\times10\%$ & $10\%\times5\%$ & $\sigma^2=10$ & $\sigma^2=20$ & $\sigma^2=30$ \\
    \midrule
    \multirow{4}{*}{\rotatebox{90}{Milan}} 
      & Acc.   & \textbf{89.5$\pm$0.6} & 88.3$\pm$1.9 & 88.3$\pm$1.9 & 88.3$\pm$1.9 & 89.7$\pm$2.0 & 89.7$\pm$2.0 & 89.5$\pm$1.9 \\
      & Prec.  & \textbf{89.7$\pm$0.5} & 88.5$\pm$1.9 & 88.4$\pm$1.9 & 88.5$\pm$2.0 & 89.5$\pm$2.1 & 89.5$\pm$2.1 &89.3$\pm$2.0 \\
      & Rec.    & \textbf{89.5$\pm$0.6} & 88.3$\pm$1.9 & 88.3$\pm$1.9 & 88.3$\pm$1.9 & 89.7$\pm$2.0 & 89.7$\pm$2.0 & 89.5$\pm$1.9 \\
      & F1   & \textbf{89.5$\pm$0.5} & 87.8$\pm$1.9 & 87.7$\pm$2.0 & 87.8$\pm$2.0 & 89.4$\pm$2.1 & 89.4$\pm$2.1 & 89.2$\pm$2.0 \\
    \midrule
    \multirow{4}{*}{\rotatebox{90}{Cairo}}
      & Acc.   & \textbf{88.9$\pm$1.3} & 86.6$\pm$1.7 & 86.4$\pm$1.6 & 86.2$\pm$1.4 & 86.3$\pm$1.6 & 86.3$\pm$1.6 & 86.3$\pm$1.6 \\
      & Prec. & \textbf{88.8$\pm$1.9} & 88.3$\pm$2.1 & 88.2$\pm$2.0 & 87.8$\pm$1.8 & 88.1$\pm$2.1 & 88.1$\pm$2.1 & 88.1$\pm$2.1 \\
      & Rec.    & \textbf{88.9$\pm$1.3} & 86.6$\pm$1.7 & 86.4$\pm$1.6 & 86.2$\pm$1.4 & 86.3$\pm$1.6 & 86.3$\pm$1.6 & 86.3$\pm$1.6 \\
      & F1   & \textbf{88.7$\pm$1.6} & 86.7$\pm$1.8 & 86.6$\pm$1.7 & 86.3$\pm$1.5 & 86.5$\pm$1.7 & 86.5$\pm$1.7 & 86.5$\pm$1.7 \\
    \midrule
    \multirow{4}{*}{\rotatebox{90}{Kyoto7}} 
      & Acc.   & \textbf{73.3$\pm$2.6} & 73.7$\pm$4.2 & 74.1$\pm$4.0 & 73.7$\pm$4.2 & 73.4$\pm$4.5 & 73.0$\pm$4.8 & 73.4$\pm$4.5 \\
      & Prec. & \textbf{74.9$\pm$1.7} & 74.1$\pm$4.1 & 74.6$\pm$3.9 & 74.1$\pm$4.1 & 74.1$\pm$3.9 & 73.9$\pm$4.0 & 74.1$\pm$3.9\\
      & Rec.   &\textbf{73.8$\pm$2.5} & 73.7$\pm$4.2 & 74.1$\pm$4.0 & 73.7$\pm$4.2 & 73.4$\pm$4.5 & 73.0$\pm$4.8 & 73.4$\pm$4.5 \\
      & F1   & \textbf{73.3$\pm$2.6} & 73.1$\pm$4.2 & 73.5$\pm$3.9 & 73.1$\pm$4.2 & 72.8$\pm$4.4 & 72.5$\pm$4.7 & 72.8$\pm$4.4 \\
    \bottomrule
  \end{tabular}}}
  \label{tab:rob}
\end{table}

\begin{table}[tb]
  \centering
    \caption{{DeepCASAS and TDOST with Corrupt($5\%\times5\%$)}}
  \resizebox{\columnwidth}{!}{
    \setlength{\tabcolsep}{1mm}{
  \begin{tabular}{ccc cccc}
    \toprule
    \multirow{2}{*}{Baseline} & \multicolumn{2}{c}{Milan}&\multicolumn{2}{c}{Cairo}&\multicolumn{2}{c}{Kyoto7}\\
    \cmidrule(l){2-3} \cmidrule(l){4-5} \cmidrule(l){6-7} 
       & {Accuracy} & {F1-Score} & {Accuracy} & {F1-Score} & {Accuracy} & {F1-Score} \\
    \midrule
    DeepCASAS &  84.2$\pm$0.4 ($\downarrow 1.1\%$) & 83.4$\pm$0.9 ($\downarrow 1.4\%$)  & 72.7$\pm$1.6 ($\downarrow 7.5\%$) & 72.2$\pm$1.2 ($\downarrow 6.7\%$)  & 60.2$\pm$2.8 ($\downarrow 2.1\%$) & 59.2$\pm$3.3 ($\downarrow 1.1\%$)\\
    TDOST &  85.0$\pm$0.4 ($\downarrow 3.7\%$) & 84.3$\pm$0.4 ($\downarrow 4\%$) & 71.7$\pm$1.9 ($\downarrow 9.3\%$) &  70.1$\pm$2.1 ($\downarrow 9.2\%$) & 72.0$\pm$2.0 ($\uparrow 2\%$) & 70.5 $\pm$1.3 ($\uparrow 0.5\%$)\\
  \bottomrule
  \end{tabular}}}
  \label{tab:TDOST_corrupt}
\end{table}

\noindent \textbf{Results.} As shown in Tab.~\ref{tab:rob}, our BiL–Res model exhibits strong resilience: accuracy on \textit{Milan}, \textit{Cairo} and \textit{Kyoto7} drops by only 1--3\% across all corruption settings. {We further conducted robustness evaluations of DeepCASAS and TDOST under the $5\%\times 5\%$ corruption in Tab.~\ref{tab:TDOST_corrupt}. Overall, CARE is consistently robust across datasets and degrades less than strong baselines under the same corruption setting.}

We believe the robustness arises from three factors: (1) our preprocessing pipeline filters unstable signals and low-activation sensors, suppressing noise before encoding; (2) the dual-branch sequence--image architecture provides redundancy, allowing one representation view to compensate when the other is degraded; and (3) the SICA loss enforces consistency across views and samples, acting as a strong regularizer against distribution shifts.
\begin{table}[tb]
\centering
\caption{Performance Comparison under Time-aware Split}
\resizebox{1\columnwidth}{!}{
\setlength{\tabcolsep}{1mm}{
\begin{tabular}{cccc ccc}
\toprule
\multirow{2}{*}{Model} & \multicolumn{2}{c}{Milan} & \multicolumn{2}{c}{Cairo} & \multicolumn{2}{c}{Kyoto7} \\
\cmidrule(lr){2-3} \cmidrule(lr){4-5} \cmidrule(lr){6-7}
 &{Accuracy}  & {F1-Score}& {Accuracy} & {F1-Score} & {Accuracy} & {F1-Score} \\
\midrule
DeepCASAS (LSTM) & $74.7 \pm 0.9$  & $71.0 \pm 1.0$ & $63.9 \pm 2.0$ & $55.5 \pm 3.0$ & $51.6 \pm 7.6$ & $48.0 \pm 9.2$ \\
DeepCASAS (BiLSTM) & $75.7 \pm 0.5$ & $72.5 \pm 0.6$ & $68.7 \pm 4.8$ & $62.8 \pm 7.5$ & $55.1 \pm 1.2$ & $51.4 \pm 1.7$ \\
LSTM-ResNet18 &
\textbf{80.3 $\pm$ 1.3} & \textbf{78.2 $\pm$ 1.5} & \textbf{73.9 $\pm$ 2.5} &
\textbf{72.6 $\pm$ 3.5} & \textbf{68.6 $\pm$ 1.9} & \textbf{67.1 $\pm$ 2.9} \\
BiLSTM-ResNet18 &
\underline{$77.9 \pm 0.4$} & \underline{$75.8 \pm 1.1$} & \underline{$73.2 \pm 3.2$} &
\underline{$70.7 \pm 2.0$} & \underline{$68.1 \pm 4.5$} & \underline{$65.3 \pm 6.8$} \\

\bottomrule
\end{tabular}}}
\label{tab:timeaware}
\end{table}

{Tab.~\ref{tab:timeaware} reports results under the time-aware split. Despite temporal-drift degradation for both methods, ours consistently outperforms DeepCASAS by about $2\%$ on \textit{Milan}, $5\%$ on \textit{Cairo}, and $13\%$ on \textit{Kyoto7}.}

\section{Conclusion}
In this paper, we presented CARE, a unified framework that bridges the gap between sequence- and image-based encodings for event-triggered sensor data. Our results demonstrate that explicitly aligning temporal and spatial cues through supervised contrastive learning not only advances performance beyond unimodal and naïve fusion baselines but also improves robustness to common real-world challenges such as sensor failures and layout variability. Importantly, CARE achieves this with a one-stage joint training strategy, making it more deployment-ready than multi-stage pipelines.

Beyond state-of-the-art performance, our findings highlight a broader insight: temporal and spatial representations are not merely complementary but mutually reinforcing when aligned in a shared latent space. This opens promising directions for generalizing contrastive alignment to other multimodal sensing domains (e.g., combining audio, physiological, or textual streams) and for developing adaptive mechanisms that handle evolving sensor infrastructures in long-term smart home deployments. By providing both methodological advances and practical robustness, CARE represents a step toward more reliable and scalable ambient intelligence systems.

{However, our model still has some limitations. First, we follow the standard CASAS pre-segmentation protocol, which may limit deployment on continuous streams; in practice, CARE can be used after a lightweight segmentation front-end. Second, extending CARE to multi-inhabited homes remains challenging, as overlapping activities yield entangled event streams and complex spatiotemporal dependencies. Handling overlap-aware recognition is a key direction for future work. Third, SICA uses labels to construct class-level contrastive pairs, which may be difficult to obtain in privacy-sensitive smart homes, motivating more label-efficient variants. Finally, the spatial image relies on fixed sensor coordinates, which may not directly generalize to reconfigurable environments. The dual-encoder design also increases memory usage, motivating lighter backbones and layout-adaptive variants in future work. }

\newpage
\bibliographystyle{plain}
\bibliography{kdd24.bib}

\end{document}